\documentclass{article}

\usepackage{microtype}
\usepackage{graphicx}
\usepackage{subcaption}
\usepackage{booktabs} 

\usepackage{enumitem}
\usepackage{hyperref}
\usepackage{url}
\usepackage{wrapfig}
\usepackage{graphicx}
\usepackage{tikz}
\usepackage{pifont}
\usepackage{booktabs}
\usepackage{bm}

\usepackage[preprint]{icml2026}

\usepackage{amsmath}
\usepackage{amssymb}
\usepackage{mathtools}
\usepackage{amsthm}

\definecolor{myorange}{RGB}{80, 101, 142}
\definecolor{lightblue}{RGB}{91, 182, 220}
\definecolor{bblue}{RGB}{51,102,204}
\definecolor{myred}{RGB}{204,0,0} 
\hypersetup{
    colorlinks=true,
    linkcolor=red,
    filecolor=myorange,
    urlcolor=lightblue,
    citecolor=lightblue,
}

\definecolor{red}{RGB}{192,0,0}
\definecolor{blue}{RGB}{48, 85, 151}

\usepackage{xurl}
\usepackage{amsfonts}       
\usepackage{nicefrac}       
\usepackage[normalem]{ulem}
\useunder{\uline}{\ul}{}
\usepackage{color}    
\usepackage{caption}
\usepackage[table]{xcolor}
\usepackage{tcolorbox}
\usepackage[capitalize,noabbrev]{cleveref}

\theoremstyle{plain}

\theoremstyle{definition}

\theoremstyle{remark}

\usepackage[textsize=tiny]{todonotes}

\icmltitlerunning{ToR: Token-Reweighting for RLVR in Multimodal LLMs}

\begin{document}

\twocolumn[
  \icmltitle{Bridging Perception and Reasoning: \\ Token Reweighting for RLVR in Multimodal LLMs}



  \icmlsetsymbol{equal}{*}

  \begin{icmlauthorlist}
    \icmlauthor{Jinda Lu}{ustc}
    \icmlauthor{Junkang Wu}{ustc}
    \icmlauthor{Jinghan Li}{ustc}
    \icmlauthor{Kexin Huang}{ustc}
    \icmlauthor{Shuo Yang}{pku}  \\
    \icmlauthor{Guoyin Wang}{indepent}
    \icmlauthor{Jiancan Wu}{ustc}
    \icmlauthor{Xiang Wang}{ustc}
    \icmlauthor{Xiangnan He}{ustc}
  \end{icmlauthorlist}

  \icmlaffiliation{ustc}{University of Science and Technology of China}
  \icmlaffiliation{pku}{Peking University}
  \icmlaffiliation{indepent}{Independent Researcher}

  \icmlcorrespondingauthor{Jinda Lu}{lujd@mail.ustc.edu.cn}

  \icmlkeywords{MLLM, RL}

  \vskip 0.3in
]



\printAffiliationsAndNotice{}  

\begin{abstract}
Extending Reinforcement Learning with Verifiable Rewards (RLVR) to multimodal large language models (MLLMs) faces a fundamental challenge: their responses inherently interleave \textbf{perception-related tokens}, which ground visual content, with \textbf{reasoning-related tokens}, which construct reasoning chains. These token types instantiate distinct yet interdependent capacities --- \emph{visual grounding and symbolic reasoning} --- making isolated optimization insufficient.
Through token-level empirical analysis, we demonstrate that optimizing either perception- or reasoning-only tokens consistently underperforms full optimization, underscoring their inherent coupling.
To address this, we propose a \emph{plug-and-play} \textbf{To}ken-\textbf{R}eweighting (\textbf{ToR}) strategy that explicitly models this interdependence by identifying critical tokens of both types and dynamically reweighting them during RLVR training.
Applied on top of existing methods (\textit{e.g.}, GRPO and DAPO), ToR delivers consistent performance gains across multiple multi-modal reasoning benchmarks, achieving state-of-the-art performance with both accurate visual grounding and coherent reasoning.
\end{abstract}
\section{Introduction}
\begin{figure}[t]
    \begin{center}
\centerline{\includegraphics[width=0.47\textwidth]{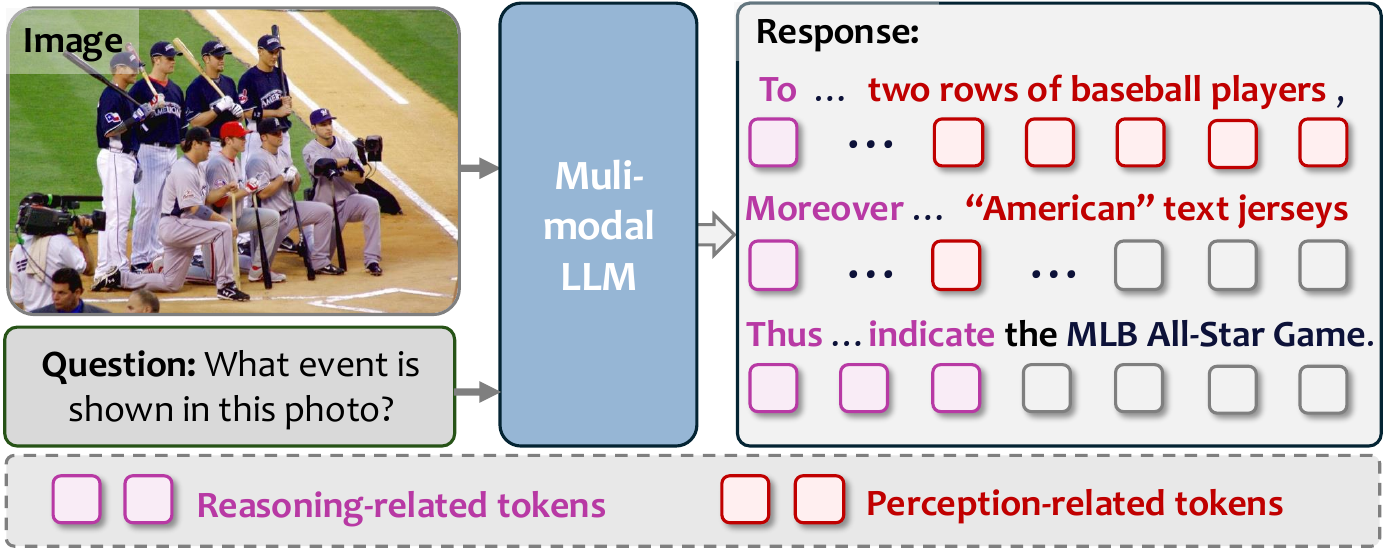}}
    \caption{MLLM responses typically involve two types of critical tokens: \textbf{(1)} reasoning-related tokens to construct reasoning chains, and \textbf{(2)} perception-related tokens to ground visual content.}
    \vskip -0.4in
    \label{fig1:teaser}
    \end{center}
\end{figure}

Reinforcement Learning with Verifiable Rewards (RLVR) has substantially advanced the reasoning ability of large language models (LLMs) on complex tasks~\citep{tulu3, deepseekmath, deepseek_r1, qwen3}.
Extending RLVR to multimodal large language models (MLLMs), however, is non-trivial: generated responses \emph{interleave} tokens that ground visual content (\textbf{perception}) with tokens that drive symbolic inference (\textbf{reasoning}), as illustrated in Figure~\ref{fig1:teaser}.
Existing MLLM RLVR variants typically optimize these capabilities in isolation --- either via chain-of-thought objectives for reasoning~\citep{vision_r1, cold_start} or perception-oriented augmentations for perception~\citep{PAPO, perception_r1, noisyrollout} --- leaving their interaction underexplored.

We hypothesize that this separated optimization is suboptimal because perception and reasoning are fundamentally interdependent at the token level. 
To empirically validate this claim, we conduct a controlled ``selective optimization'' study under Group Relative Policy Optimization (GRPO)~\citep{deepseekmath}.
We identify reasoning-related tokens via high \textbf{next-token entropy} (following recent insights on reasoning forks~\citep{80_20_tokens, llm_reasoning_with_exploration}), and perception-related tokens via \textbf{visual sensitivity}, measured as the change in token log-probability when conditioning on the image versus a text-only context (details in Section~\ref{sec:approach}).
We then train models while masking gradients on non-selected tokens, comparing three settings: optimizing only reasoning-related tokens, only perception-related tokens, and all tokens (vanilla GRPO).

\begin{figure*}[t]
\begin{center}
\center{\includegraphics[width=0.95\textwidth]{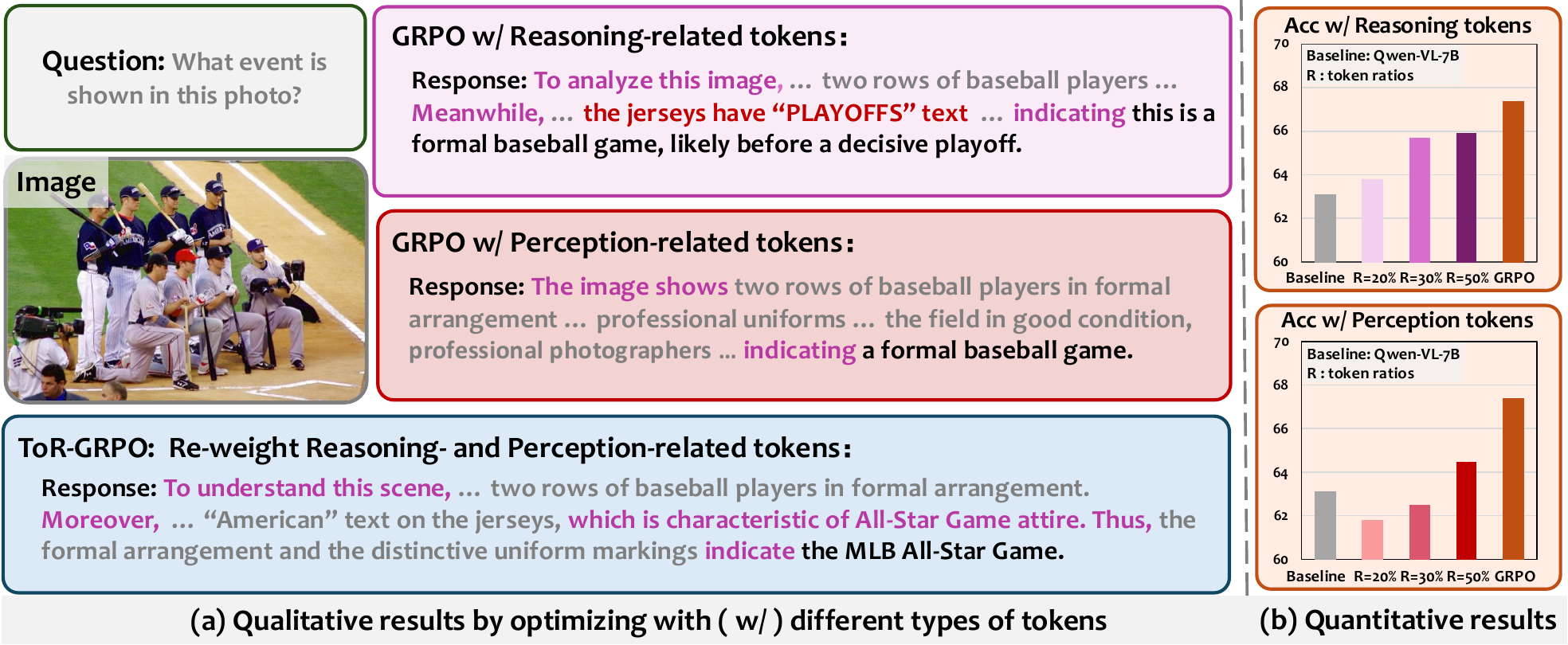}}
\end{center}
\caption{Performance comparison over the wemath benchmark~\citep{wemath} when optimizing different token types with GRPO~\citep{deepseekmath}. Results across selection ratios 20\%, 30\%, or 50\% show that optimizing either reasoning-only or perception-only tokens underperforms all tokens. 
Qualitative examples are selected from the best-performing checkpoints.}
\label{Figure-1-exp}
\end{figure*}

Across selection ratios (20\%, 30\%, 50\%), optimizing only reasoning tokens underperforms vanilla GRPO (\textit{e.g.}, $\sim$2\% absolute drop on we-math benchmark), and optimizing only perception tokens fares worse (\textit{e.g.}, $\sim$3\% drop, with 20\% and 30\% tokens even worse than the baseline model); neither matches training on all tokens (Figure~\ref{Figure-1-exp}). Qualitatively, reasoning-only models produce coherent-looking chains of thought yet misinterpret key visual content, while perception-only models preserve low-level grounding but fail to integrate it into coherent reasoning. These results support our hypothesis: \emph{perception and reasoning are coupled capabilities that demand joint optimization}.

Building on this insight, we propose \textbf{To}ken-\textbf{R}eweighting (\textbf{ToR}), a lightweight, plug-and-play module for RLVR that jointly optimizes perception and reasoning. Instead of treating all tokens equally or optimizing subsets in isolation, ToR strategically identifies the most critical perception- and reasoning-related tokens and adaptively reweights their importance in the policy gradient calculation. This mechanism explicitly models their interdependence, encouraging the MLLM to integrate visual grounding into its logical deliberations. As shown in Figure~\ref{Figure-1-exp}, applying ToR to GRPO (\textbf{ToR-GRPO}) not only recovers the performance lost from selective optimization but surpasses the standard GRPO baseline, confirming our effectiveness.

In summary, we show that perception and reasoning in MLLMs are tightly coupled at the token level and should be optimized jointly.
By explicitly reweighting perception- and reasoning-related tokens during training, Token Reweighting enables MLLMs to improve reasoning decisions while preserving accurate visual grounding, resulting in consistent gains across various models and benchmarks.
\section{Preliminaries}
In this section, we revisit Reinforcement Learning with Verifiable Rewards (RLVR) procedure and representative RLVR optimization strategies for Multi-modal Large Language Models (MLLMs). 

\subsection{Reinforcement Learning with Verifiable Rewards}
Reinforcement Learning with Verifiable Rewards (RLVR) enhances multi-modal large language models (MLLMs) by aligning model outputs with verifiable answers. Given a multi-modal input with ground truth from a batch of B samples, $ (\mathrm{I}^{b}, \mathrm{q}^{b})) \in \{\mathrm{I}^b, \mathrm{q}^b\}_{b=1}^{B}$, and $y^{b} \in \{ y^{b}\}_{b=1}^{B}$, where $\mathrm{I}^{b}$, $\mathrm{q}^{b}$, and $y^{b}$ denotes the image, question, and ground-truth respectively, the model $\pi_{\theta}$ generates output $\bm{o}^{b}$ containing a reasoning process and a prediction. The prediction is enclosed in \texttt{\textbackslash boxed\{.\}} while reasoning is delimited by \texttt{<think>...</think>}, enabling automated verification against the ground truth answers.
RLVR employs a binary reward function $\mathbf{R}(\cdot)$ to determine whether the answer is correct by comparing the model output $\bm{o}^{b}$ with ground truth $y^{b}$.
The goal of RLVR is to maximize the reward function, formalized as:
\begin{equation}
\resizebox{.91\hsize}{!}{
$\mathcal{J}_{\mathrm{RLVR}}(\theta) = 
\displaystyle \max_{\theta} 
\mathbb{E}_{\{(\mathrm{I}^b, \mathrm{q}^{b}) \mid y^{b}) \}_{b=1}^{B}}
\mathbb{E}_{\bm{o}^{b} \sim \pi_{\theta}(\cdot \mid \mathrm{I}^{b}, \mathrm{q}^{b})}[
\mathbf{R}(\bm{o}^{b}, y^{b})].$
}
\end{equation}

\subsection{RLVR Optimization Algorithms}

\textbf{Group Relative Policy Optimization (GRPO).} As a widely adopted RLVR optimization strategy, GRPO stabilizes training by computing advantages within response groups~\citep{deepseekmath}. Concretely, given a batch of samples $\{(\mathrm{I}^b, \mathrm{q}^b) \mid y^{b}\}_{b=1}^{B}$, the GRPO objective is:
\begin{equation}
\resizebox{.91\hsize}{!}{
$\begin{aligned}
\mathcal{J}_{\text{GRPO}}&(\theta)  = \mathbb{E}_{\substack{q \sim P(Q) \\ \{o_i\} \sim \pi_{\theta_{\text{old}}}}} \Bigg[ \frac{1}{G} \sum_{i=1}^G \frac{1}{|o_i|} \sum_{t=1}^{|o_i|} \min \bigg( \frac{\pi_\theta(o_{i,t})}{\pi_{\theta_{\text{old}}}(o_{i,t})} \hat{A}_{i,t}, \\
&\text{clip} \left( \frac{\pi_\theta(o_{i,t})}{\pi_{\theta_{\text{old}}}(o_{i,t})}, 1 \pm \epsilon \right) \hat{A}_{i,t} \bigg) - \beta \cdot \mathbb{D}_{KL}(\pi_\theta \| \pi_{\text{ref}}) \Bigg],
\end{aligned}$}
\end{equation}
where $G$ is the rollout group size for each input, and the clip function restricts the importance ratio within $[1 - \epsilon, 1 + \epsilon]$. Moreover, the advantage can be formulated as:
\begin{equation}
\scalebox{0.9}{
$\begin{cases}
    \hat{A}^{b}_{i,t} = \dfrac{\mathbf{R}^{b}_i - \text{mean}(\mathbf{R}^{b})}{\text{std}(\mathbf{R}^{b})}, \\[10pt]
    \mathbf{R}^{b}_{i} = \mathbb{I}(\text{is\_equivalent}(o^{b}_{i}, y^{b})),
\end{cases}$
}
\end{equation}
where $\mathbb{I}(\cdot)$ is the indicator function, \texttt{is\_equivalent}$(\cdot)$ extracts the predictions from the model output $o^{b}_{i}$ and compares with the ground truth $y^{b}$.

\textbf{Decoupled Clip and Dynamic Sampling Policy Optimization (DAPO)}.
DAPO~\citep{dapo} improves upon GRPO by removing KL regularization and introducing clip-higher, dynamic sampling, and token-level loss from the GRPO loss, achieving new state-of-the-art performance.
Specifically, the DAPO objective for the batch of samples $\{(\mathrm{I}^b, \mathrm{q}^b) \mid y^{b}\}_{b=1}^{B}$ can be formalized as:
\begin{equation}
\resizebox{.91\hsize}{!}{
$\begin{aligned}
\mathcal{J}_{\text{DAPO}}&(\theta) = \mathbb{E}_{\substack{q \sim P(Q) \\ \{o_i\} \sim \pi_{\theta_{\text{old}}}}} \Bigg[ \frac{1}{\sum |o_i|} \sum_{i=1}^{G} \sum_{t=1}^{|o_i|} \min \bigg( \frac{\pi_\theta(o_{i,t})}{\pi_{\theta_{\text{old}}}(o_{i,t})} \hat{A}_{i,t}, \\
& \text{clip} \left( \frac{\pi_\theta(o_{i,t})}{\pi_{\theta_{\text{old}}}(o_{i,t})}, 1 - \epsilon_{\text{low}}, 1 + \epsilon_{\text{high}} \right) \hat{A}_{i,t} \bigg) \Bigg],
\end{aligned}$}
\end{equation}
where $\epsilon_{\text{low}}$ and $\epsilon_{\text{high}}$ decouple the clipping thresholds for negative and positive advantages, respectively. In this work, we apply our token-reweighting strategy to both GRPO and DAPO, demonstrating its general applicability across various RLVR optimization strategies.
\section{Approach}
\label{sec:approach}

In this section, we introduce \textbf{Token Reweighting (ToR)}, a simple strategy that jointly optimizes reasoning- and perception-related tokens for reinforcement learning.
We first describe how to identify these two types of critical tokens, and then present how ToR integrates them into existing RLVR objectives such as GRPO and DAPO.

\subsection{Token Identification}
Multimodal reasoning responses interleave tokens that serve distinct functional roles.
Some tokens correspond to critical reasoning decisions, while others primarily ground the response in visual inputs.
We therefore identify two complementary token types---\emph{reasoning-related} and \emph{perception-related} tokens---using the intrinsic model signals.

\subsubsection{Reasoning-related tokens}

\textbf{Identification.}
Recent studies~\citep{80_20_tokens, llm_reasoning_with_exploration} have shown that tokens with high predictive entropy often correspond to pivotal ``forking'' points in reasoning chains, reflecting the model’s decision uncertainty.
Motivated by this observation, we identify reasoning-related tokens based on token-level entropy during rollout generation.
Given the $i$-th generated response $\mathbf{o}^{b}_{i} = \{o^{b}_{i,1}, o^{b}_{i,2} \ldots, o^{b}_{i,L_i}\}$ conditioned on the image $\mathrm{I}^{b}$, and then question $\mathrm{q}^{b}$ from the input batch $\{(\mathrm{I}^{b}, \mathrm{q}^{b})\}_{b=1}^{B}$, the prediction entropy at position $t$ is computed as:
\begin{equation}
\begin{aligned}
\label{eq:reasoning_entropy}
H^{b}_{i,t} = & -\sum_{v \in \mathcal{V}_{\text{top-}p}} 
\mathrm{P}_\theta(o^{b}_{i,t} = v \mid \mathbf{o}^{b}_{i,<t}, \mathrm{I}^{b}, \mathrm{q}^{b}) \\
& \cdot \log \mathrm{P}_\theta(o^{b}_{i,t} = v \mid \mathbf{o}^{b}_{i,<t}, \mathrm{I}^{b}_{i}, \mathrm{q}^{b}_{i}),
\end{aligned}
\end{equation}
where $\mathcal{V}_{\text{top-}p}$ denotes the set of vocabulary tokens within the top-$p$ cumulative probability ($p=0.95$). This top-$p$ truncation avoids the influence of extremely low-probability tokens and is consistent with the rollout sampling process.

To aggregate across the rollout batch, we collect all token entropies into a set:
\begin{equation}
\resizebox{.91\hsize}{!}{
$\mathcal{H} = \{ H^{b}_{i,t} \mid b=1,\ldots, B;\; i=1,\ldots,G; \; t=1,\ldots,L^{b}_i \}.$}
\end{equation}
The reasoning-token set is then defined by selecting the top-$\alpha_r$ fraction of tokens with the highest entropy across the batch:
\begin{equation}
\label{eq:reasoning_tokens_final}
\mathcal{T}_{\mathrm{r}}
= \{ (b,i,t) \mid H^{b}_{i,t} \geq \operatorname{Percentile}_{1-\alpha_r}(\mathcal{H}) \},
\end{equation}
where $\alpha_r$ controls the fraction of selected tokens. Intuitively, these high-uncertainty positions correspond to pivotal points where reasoning chains are shaped.

\begin{figure*}[t]
\centering
\includegraphics[width=0.8\textwidth]{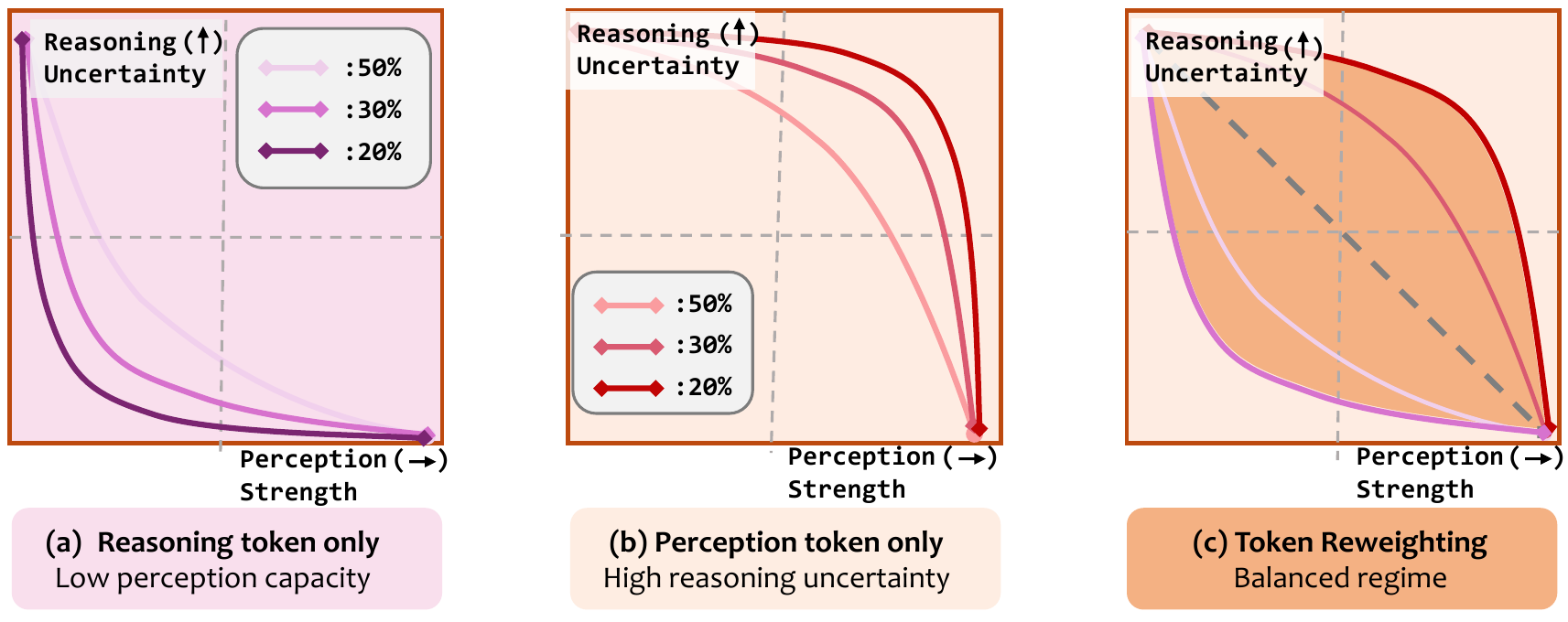}
\caption{\textbf{Comparison of perception strength and reasoning uncertainty under different token ratios during GRPO optimization.}
(a) Reasoning-token-only optimization suffers from limited perception capacity.
(b) Perception-token-only optimization is sensitive to high reasoning uncertainty.
(c) Token reweighting yields a balanced regime by adaptively trading off perception strength and reasoning uncertainty, where the shaded region indicates effective optimization outcomes.
Dashed lines denote the balance where increased perception strength compensates for reasoning uncertainty.}
\label{fig:method_summary}
\end{figure*}

\textbf{Influences of reasoning-related tokens.} 
Restricting optimization to the reasoning-token set $\mathcal{T}_{\mathrm{r}}$ concentrates gradient updates on high-entropy decision points, thereby reducing reasoning uncertainty, as illustrated in Figure~\ref{fig:method_summary}(a).
However, excluding perception-related tokens weakens visual grounding, allowing perception errors to persist during training.
As a result, reasoning-only optimization sharpens local decision confidence but fails to support robust multimodal reasoning, leading to inferior performance compared to full-token GRPO.

\subsubsection{Perception-related tokens}

\textbf{Identification of perception-related tokens.} 
While reasoning-related tokens capture decision uncertainty, perception-related tokens highlight positions whose predictions strongly depend on visual inputs. 
To quantify this dependence, we compare the token log-probabilities under two conditions: (i) \emph{with image}, conditioned on both $\mathrm{I}^{b}$ and $\mathrm{q}^{b}$; and (ii) \emph{without image}, where the image channel is replaced by an empty placeholder $\bm{\varnothing}$ and the model is conditioned only on $\mathrm{q}^{b}$. 

Given the $i$-th generated response $\mathbf{o}^{b}_{i} = \{o^{b}_{i,1}, \ldots, o^{b}_{i,T_i}\}$, the visual-sensitivity score at position $t$ is computed as:
\begin{equation}
\label{eq:perception_logdiff}
\resizebox{.91\hsize}{!}{
$S^{b}_{i,t} = \big| \log \pi_{\theta}(o^{b}_{i,t} \mid \mathbf{o}^{b}_{i,<t}, \mathrm{I}^{b}_{i}, \mathrm{q}^{b}_{i}) - \log \pi_{\theta}(o^{b}_{i,t} \mid \mathbf{o}^{b}_{i,<t}, \bm{\varnothing}, \mathbf{q}^{b}_{i}) \big|,
$}
\end{equation}
where $\pi_{\theta}(o_t\mid \cdot)$ is the token-level policy. A large $S_{i,t}$ indicates a strong visual influence on token $o_{i,t}$.
Aggregating across the rollout batch, the perception-token set is defined as the top-\(\alpha_p\) fraction of tokens with the highest visual-sensitivity:
\begin{equation}
\label{eq:perception_tokens_final}
\resizebox{.91\hsize}{!}{$
\begin{aligned}
S & = \{S^{b}_{i,t} \mid b=1,\dots,B;\; i=1 ,\ldots,G; \; t=1,\dots,L^{b}_i\} \\
\mathcal{T}_{\mathrm{p}} & =  \{ (b,i,t) \mid S^{b}_{i,t} \ge \operatorname{Percentile}_{1-\alpha_p}(S) \},
\end{aligned}$}
\end{equation}

where $\alpha_p$ controls the selection ratio. This batch-level percentile selection ensures consistent token selection thresholds across rollout batches.

\textbf{Influences of perception-related tokens.}
Restricting optimization to the perception-token set $\mathcal{T}_{\mathrm{p}}$ concentrates gradient updates on visually sensitive positions, thereby strengthening visual grounding, as conceptually illustrated in Figure~\ref{fig:method_summary}(b).
However, excluding reasoning-related tokens prevents effective optimization of high-uncertainty decision points, leaving reasoning ambiguity unresolved.
As a result, perception-only optimization preserves low-level visual fidelity but fails to support coherent reasoning, leading to inferior performance compared to full-token GRPO.

\subsection{Token Reweighting}
As conceptually illustrated in Figure~\ref{fig:method_summary}, optimizing only reasoning-related or perception-related tokens leads to imbalanced optimization.
Reasoning-only optimization reduces uncertainty but lacks visual grounding, while perception-only optimization improves grounding without resolving high-uncertainty reasoning decisions.

Motivated by this observation, we introduce \emph{Token Reweighting} (ToR), a unified framework that jointly incorporates reasoning- and perception-critical tokens.
Instead of masking gradients to a single token subset, ToR assigns token-specific weights during policy optimization, enabling simultaneous reduction of reasoning uncertainty and strengthening of visual grounding.
In this way, ToR explicitly models the interdependence between reasoning and perception at the token level.

Specifically, for the given input batch $\{(\mathrm{I}^{b}, \mathrm{q}^{b})\}_{b=1}^{B}$, we constructed the reasoning-related token set $\mathcal{T}_{\text{r}}$ and the perception-related token set $\mathcal{T}_{\text{p}}$, and thus the GRPO objective with token reweighting (\textbf{ToR-GRPO}) is:
\begin{equation}
\resizebox{.91\hsize}{!}{$
\begin{aligned}
\mathcal{J}&_{\text{ToR-GRPO}}(\theta) = \ \mathbb{E}_{\{(\mathrm{I}^b, \mathrm{q}^{b})\}_{b=1}^{B}}\mathbb{E}_{\{\bm{o}^{b}_i\}_{i=1}^G \sim \pi_{\theta_{\text{old}}}} \Bigg\{ \frac{1}{G} \sum_{i=1}^G \frac{1}{|o^{b}_i|} \sum_{t=1}^{|o^{b}_i|} \\
& \Big[ \gamma_{\text{r}} \cdot \mathbb{I}[(b,i,t) \in \mathcal{T}_{\text{r}}] + \gamma_{\text{p}} \cdot \mathbb{I}[(b,i,t) \in \mathcal{T}_{\text{p}}] \Big] \cdot \min \Big[ r_\theta(o^{b}_{i,t}) \hat A^{b}_{i,t}, \\
& \text{clip} ( r_\theta(o^{b}_{i,t}), 1-\epsilon, 1+\epsilon ) \hat A^{b}_{i,t} \Big] - \beta \mathbb{D}_{KL}[\pi_\theta \,\|\, \pi_{\text{ref}}] \Bigg\}
\end{aligned}$}
\end{equation}

Similarly, the DAPO objective with the token-weighting strategy (\textbf{ToR-DAPO}) is:
\begin{equation}
\resizebox{.91\hsize}{!}{$
\begin{aligned}
\mathcal{J}&_{\text{ToR-DAPO}}(\theta) = \ \mathbb{E}_{\{(\mathrm{I}^b, \mathrm{q}^{b})\}_{b=1}^{B}}\mathbb{E}_{\{\bm{o}^{b}_i\}_{i=1}^G \sim \pi_{\theta_{\text{old}}}} \left\{ \frac{1}{\sum_{i=1}^{G} |o^{b}_i|} \right. \\
& \sum_{i=1}^{G} \sum_{t=1}^{|o_i|} \Big[ \gamma_{\text{r}} \cdot \mathbb{I}[(b,i,t) \in \mathcal{T}_{\text{r}}] + \gamma_{\text{p}} \cdot \mathbb{I}[(b,i,t) \in \mathcal{T}_{\text{p}}] \Big], \\ 
&\left. \cdot \min \Big[ r_\theta(o^{b}_{i,t}) \hat A^{b}_{i,t}\,, \text{clip} ( r_\theta(o^{b}_{i,t}), 1-\epsilon_{\text{low}}, 1+\epsilon_{\text{high}} ) \hat A^{b}_{i,t} \Big] \right\}.
\end{aligned}$}
\end{equation}
$r_\theta(o^{b}_{i,t})$ is the importance sampling ratio, formalized as:   
\begin{equation}
r_\theta(o^{b}_{i,t}) = \frac{\pi_\theta(o^{b}_{i,t} \mid \mathrm{I}^{b}_{i},\mathrm{q}^{b}_{i},o^{b}_{i,<t})}{\pi_{\theta_{\text{old}}}(o^{b}_{i,t} \mid \mathrm{I}^{b}_{i},\mathrm{q}^{b}_{i},o^{b}_{i,<t})}.
\end{equation}

Here, $\gamma_{\text{r}}$ weights reasoning-related tokens, emphasizing critical decision points in reasoning chains; $\gamma_{\text{p}}$ weights perception-related tokens, emphasizing the integration of visual context. Tokens outside these sets are excluded from optimization (\textit{i.e,} assigned a weight of zero for advantage calculation). 
This reweighting ensures gradients focus on tokens essential for both reasoning and perception, improving both training efficiency and effectiveness.

\textbf{Discussion}
Compared with existing approaches, ToR offers several key advantages: 
\ding{182} \textbf{Plug-and-play}. A simple reweighting mask integrates seamlessly into standard RLVR objectives, without introducing extra pipeline modifications.
\ding{183} \textbf{Self-contained.} Critical tokens are identified purely from the model’s intrinsic uncertainty and visual sensitivity, eliminating the need for external priors.
\ding{184} \textbf{Joint optimization.} By explicitly modeling the interdependence between reasoning and perception tokens, ToR enables balanced and simultaneous enhancement of both capabilities.
\section{Experiment}
\label{sec:experiment}
In this section, we elaborate on the effectiveness of our token reweighting strategy. Specifically, we first introduce the details of our experimental settings. Next, we present the ablation studies, and finally, we compare our results with those of state-of-the-art methods across various benchmarks. 

\subsection{Experimental Settings}
We adopt the Geometry3K~\citep{geometry3k} dataset for training and validation, following existing methods~\citep{noisyrollout, perception_r1}, which consists of 2,100 training samples and 300 validation samples. 
Moreover, we employ the multi-modal framework EasyR1~\citep{easyr1} for reinforcement learning training, and following the works in~\citep{noisyrollout} for evaluation.
Evaluation is conducted with five benchmarks, including four for visual reasoning: MathVerse~\citep{mathverse}, MathVision~\citep{mathvision}, MathVista~\citep{mathvista}, and WeMath~\citep{wemath}, and one for visual perception: HalluBench~\citep{hallusionbench}. 

\textbf{Implementation details.} 
We adopt Qwen2.5-VL-7B~\citep{qwen2_5vl} as our baseline model by following existing works in~\citep{Mm-eureka, noisyrollout}. Specifically, all experiments are conducted using 8 NVIDIA H800 GPUs (80 GB memory for each), with the default settings in EasyR1: a learning rate of $1e^{-6}$, a global batch size of 128, a rollout batch size of 512, a rollout $n$ as 12.

\subsection{Ablation Studies}
\label{sec:ablation}
We conduct ablation studies to analyze how different token-level optimization strategies affect multimodal reasoning performance.
In particular, we focus on understanding whether reasoning-related or perception-related tokens can be optimized in isolation, and whether jointly reweighting both types of tokens is necessary for robust learning.
All ablation experiments are conducted using Qwen-2.5-VL-7B with GRPO on the Geometry3K training set, and evaluated on HalluBench, WeMath, MathVerse, MathVision, and MathVista.

\paragraph{Failure of isolated token optimization.}
We first investigate whether selectively optimizing a single token type is sufficient for multimodal reasoning.
Specifically, we isolate reasoning-related tokens or perception-related tokens according to our token identification criteria, vary their proportions, and mask gradients on all remaining tokens.

Table~\ref{tab:Exp_entropy} reports results when varying the proportion of reasoning-related tokens ($\alpha_r$).
Across all benchmarks, selectively optimizing reasoning tokens consistently underperforms full-token GRPO.
Notably, even on reasoning-intensive benchmarks such as WeMath and MathVerse, reasoning-only optimization fails to match the performance of full-token training.
This indicates that effective reasoning optimization critically depends on perception-related context, and that reasoning signals alone are insufficient to guide robust policy learning in multimodal settings.

\begin{table}[t]
    \centering
    \caption{
    \textbf{Performance comparison with different ratios of reasoning-related tokens ($\alpha_{r}$).}
    We vary the proportion of reasoning-related tokens from 20\% to 80\%.
    }
    \label{tab:Exp_entropy}
    \setlength{\tabcolsep}{9pt}
    \renewcommand{\arraystretch}{1.2}
    \resizebox{\linewidth}{!}{
    \begin{tabular}{c c c c c c}
        \toprule
        $\alpha_{r}$
        & \textbf{HalluBench}
        & \textbf{MathVerse}
        & \textbf{WeMath}
        & \textbf{MathVision}
        & \textbf{MathVista} \\
        \midrule
        20\% & 62.8 & 49.3 & 63.9 & 26.5 & 69.2 \\
        30\% & 65.1 & \uline{50.7} & 65.6 & \uline{27.0} & \uline{70.0} \\
        40\% & 66.1 & 50.1 & 65.9 & 26.3 & 69.4 \\
        50\% & 67.1 & 50.4 & 66.0 & \uline{27.0} & 68.7 \\
        60\% & 67.1 & 50.0 & 67.1 & \textbf{27.3} & \textbf{70.3} \\
        70\% & \uline{67.5} & \textbf{50.8} & \uline{67.3} & 26.7 & \uline{70.0} \\
        80\% & \textbf{68.4} & 50.3 & \textbf{67.6} & 26.8 & 69.0 \\
        \midrule
        GRPO & 69.8 & 50.8 & 67.4 & 27.3 & 70.5 \\
        \bottomrule
    \end{tabular}
    }
\end{table}

\begin{table}[t]
    \centering
    \caption{
    \textbf{Performance comparison with different ratios of perception-related tokens ($\alpha_{p}$).}
    We vary the proportion of perception-related tokens from 20\% to 80\%.
    }
    \label{tab:Exp_image_diff}
    \setlength{\tabcolsep}{9pt}
    \renewcommand{\arraystretch}{1.2}
    \resizebox{\linewidth}{!}{
    \begin{tabular}{c c c c c c}
        \toprule
        $\alpha_{p}$
        & \textbf{HalluBench}
        & \textbf{MathVerse}
        & \textbf{WeMath}
        & \textbf{MathVision}
        & \textbf{MathVista} \\
        \midrule
        20\% & 65.9 & 38.6 & 61.4 & 22.7 & 65.5 \\
        30\% & \textbf{69.3} & 44.9 & 64.8 & 24.5 & 67.8 \\
        40\% & 67.8 & 44.7 & \uline{65.9} & 25.1 & 67.4 \\
        50\% & \uline{68.3} & 39.5 & 62.9 & \textbf{27.6} & \uline{68.7} \\
        60\% & 68.1 & \uline{45.1} & 65.8 & 25.0 & 67.4 \\
        70\% & 65.1 & \textbf{45.8} & \textbf{66.8} & 25.2 & \textbf{68.9} \\
        80\% & 53.4 & \uline{45.1} & 65.8 & \uline{25.9} & 68.1 \\
        \midrule
        GRPO & 69.8 & 50.8 & 67.4 & 27.3 & 70.5 \\
        \bottomrule
    \end{tabular}
    }
\end{table}

Table~\ref{tab:Exp_image_diff} shows the results of perception-only optimization by varying the proportion of perception-related tokens ($\alpha_p$).
While perception-only reweighting achieves competitive performance on perception-oriented benchmarks such as HalluBench, it leads to substantial degradation on reasoning-demanding tasks, particularly on MathVerse and WeMath.
These results highlight a complementary failure mode: perception signals alone cannot be effectively utilized without concurrent reasoning supervision.

Together, these results demonstrate that optimizing either reasoning-related or perception-related tokens in isolation is insufficient for robust multimodal reasoning.
Although both token types are individually important, their learning dynamics are tightly coupled during optimization.

\begin{table}[t]
    \centering
    \caption{
    \textbf{Performance under different combinations of reasoning and perception tokens.}
    We vary the perception-related weight from 0.1 to 1.5 while keeping the reasoning-related weight fixed at 1.0.
    }
    \label{tab:token_ratio}
    \setlength{\tabcolsep}{9pt}
    \renewcommand{\arraystretch}{1.2}
    \resizebox{\linewidth}{!}{
    \begin{tabular}{c c c c c c}
        \toprule
        \bm{$\gamma_p$} 
        & \textbf{HalluBench} 
        & \textbf{MathVerse} 
        & \textbf{WeMath} 
        & \textbf{MathVision} 
        & \textbf{MathVista} \\
        \midrule
        0.1 & 66.5 & 50.8 & 65.5 & 27.3 & 69.7 \\
        0.3 & 69.1 & \uline{52.1} & 67.4 & \textbf{28.6} & 70.8 \\
        0.5 & \uline{71.3} & \textbf{53.0} & \textbf{68.7} & \uline{28.3} & \textbf{71.9} \\
        0.7 & 70.5 & 51.1 & \uline{68.1} & 27.5 & 70.8 \\
        0.9 & \textbf{72.2} & 50.9 & 67.0 & 27.2 & \uline{71.4} \\
        1.1 & \uline{71.3} & 49.9 & 67.3 & 26.9 & 69.3 \\
        1.3 & 70.2 & 49.9 & \uline{68.1} & 26.5 & 68.9 \\
        1.5 & 69.4 & 49.6 & 66.2 & 26.8 & 68.7 \\
        \midrule
        GRPO & 69.8 & 50.8 & 67.4 & 27.3 & 70.5 \\
        \bottomrule
    \end{tabular}
    }
    \vspace{-3ex}
\end{table}

\setlength{\tabcolsep}{3pt}
\begin{table*}[t]
    \caption{
    \textbf{Performance comparison of multi-modal LLMs under different model and data scales.}
    Following existing works~\citep{noisyrollout, perception_r1}, we highlight the data size with \textcolor{blue}{blue} and \textcolor{red}{red} for \textcolor{blue}{SFT} and \textcolor{red}{RL}, respectively.
    The best value in each column is shown in \textbf{bold}, and the second-best is \underline{underlined}.
    }
    \centering
    \resizebox{\linewidth}{!}{
        \begin{tabular}{lrccccc}
            \toprule
            \textbf{Model} & \textbf{Data Size} & \textbf{MathVerse} & \textbf{MathVision} & \textbf{MathVista} & \textbf{WeMath} & \textbf{HalluBench} \\
            \midrule
            \multicolumn{7}{l}{\textit{Open-source Models}} \\
            \midrule
            InternVL-2.5-8B~\citep{internvl2.5} & - & 39.5 & 19.7 & 64.4 & - & 67.3 \\
            InternVL-3-8B~\citep{internvl3} & - & 39.5 & 29.3 & 71.6 & 37.1 & - \\
            LLaVA-OneVision-7B~\citep{llava-ov} & - & 26.2 & - & 63.2 & - & 48.4 \\
            Qwen2.5-VL-7B-Instruct~\citep{qwen2_5vl} & - & 46.2 & 25.0 & 67.5 & 63.1 & 64.6 \\
            \midrule
            \multicolumn{7}{l}{\textit{RLVR on Qwen-2.5-VL-7B}} \\
            \midrule
            R1-VL-7B~\citep{r1_vl} & \textcolor{blue}{260K}+\textcolor{red}{10K} & 40.0 & 24.7 & 63.5 & - & - \\
            Vision-R1-7B~\citep{vision_r1} & \textcolor{blue}{200K}+\textcolor{red}{10K} & 52.4 & - & \underline{73.5} & - & - \\
            R1-OneVision-7B~\citep{r1_onevision} & \textcolor{blue}{155K}+\textcolor{red}{10K} & 46.1 & 22.5 & 63.9 & 62.1 & 65.6 \\
            OpenVLThinker-7B~\citep{openvlthinker} & \textcolor{blue}{35K}+\textcolor{red}{15K} & 48.0 & 25.0 & 71.5 & 67.8 & 70.8 \\
            MM-Eureka-Qwen-7B~\citep{Mm-eureka} & \textcolor{red}{15K} & 50.5 & 28.3 & 71.5 & 65.5 & 68.3 \\
            ThinkLite-7B-VL~\citep{thinklite} & \textcolor{red}{11K} & 50.2 & 27.6 & {72.7} & 69.2 & 71.0 \\
            VLAA-Thinker-7B~\citep{vlaa_thinking} & \textcolor{red}{25K} & 49.9 & 26.9 & 68.8 & 67.9 & 68.6 \\
            NoisyRollout-7B~\citep{noisyrollout} & \textcolor{red}{2.1K} & 53.2 & 28.5 & 72.6 & 69.6 & 72.1 \\
            \midrule
            GRPO~\citep{deepseekmath} & \textcolor{red}{2.1K (Geo3K)} & 50.8 & 27.3 & 70.5 & 67.4 & 69.8 \\
            \rowcolor{lightgray} {ToR-GRPO} & \textcolor{red}{2.1K (Geo3K)} & 53.0 & 28.6 & 71.9 & 68.9 & \underline{72.4} \\
            \midrule
            DAPO~\citep{dapo} & \textcolor{red}{2.1K (Geo3K)} & 50.6 & 26.5 & 70.3 & 69.3 & 67.9 \\
            \rowcolor{lightgray} {ToR-DAPO} & \textcolor{red}{2.1K (Geo3K)} & \underline{53.4} & \underline{27.9} & 72.6 & \underline{72.1} & 71.8 \\
            \rowcolor{lightgray} {ToR-DAPO} & \textcolor{red}{39K (ViRL-39K)} & \textbf{54.3} & \textbf{31.6} & \textbf{74.2} & \textbf{73.0} & \textbf{73.6} \\
            \midrule
            \multicolumn{7}{l}{\textit{RLVR on Qwen-2.5-VL-3B}} \\
            \midrule
            DAPO & \textcolor{red}{2.1K (Geo3K)} & 42.25 & 22.6 & 64.0 & 57.8 & 54.7 \\
            \rowcolor{lightgray} {ToR-DAPO} & \textcolor{red}{2.1K (Geo3K)} & 46.20 & 26.5 & 66.0 & 59.6 & 56.3 \\
            \rowcolor{lightgray} {ToR-DAPO} & \textcolor{red}{39K (ViRL-39K)} & 47.50 & 28.3 & 67.5 & 62.4 & 60.8 \\
            \bottomrule
        \end{tabular}
    }
    \label{tab:main_result}
\end{table*}

\vspace{-1.5ex}
\paragraph{Joint reweighting of reasoning and perception tokens.}
Motivated by the above failure modes, we next investigate whether jointly reweighting reasoning- and perception-related tokens can alleviate the limitations of isolated optimization.
Table~\ref{tab:token_ratio} reports results when fixing the weight of reasoning-related tokens to 1.0 and varying the weight of perception-related tokens ($\gamma_p$).

We observe that jointly reweighting both token types consistently improves performance over isolated optimization across all benchmarks.
In particular, moderate perception token weights lead to improved reasoning performance while preserving strong perception accuracy.
In contrast, extremely low or high perception token weights degrade performance, indicating that over-emphasizing either token type disrupts the balance of effective multimodal reasoning.

Overall, a perception token weight of $\gamma_p = 0.5$ provides a strong and stable trade-off across benchmarks, and we adopt this setting as the default configuration in subsequent experiments.
Regarding the token selection ratio, we find that increasing the proportion of selected tokens generally improves performance, but also gradually approaches full-token optimization, diminishing the effect of selective reweighting.
Lower ratios, while not optimal, still yield competitive performance without destabilizing training.
We therefore adopt 30\% as a conservative operating point that provides sufficient learning signal while preserving the benefits of selective token reweighting.
Importantly, our method does not rely on precise tuning of this ratio, and similar qualitative trends are observed under nearby settings.

\vspace{-1.5ex}
\paragraph{Discussion on overlapping tokens.}
During early stages of GRPO training on Geometry3K, approximately 12\% of tokens are identified as both reasoning- and perception-related.
As training progresses, this overlap ratio evolves dynamically.
Empirically, we observe that while performance degrades when either token type is optimized in isolation, the degradation is consistently smaller when retaining reasoning-related supervision.
Therefore, for overlapping tokens, we assign them the same weight as reasoning-related tokens during reweighting, which yields more stable optimization behavior in practice.

\subsection{Comparison with state-of-the-art approaches}
We compare our token re-weighting strategy with a broad set of state-of-the-art multi-modal LLMs, including open-source instruction-tuned models (e.g., InternVL-2.5~\citep{internvl2.5}, InternVL-3~\citep{internvl3}, LLaVA-OneVision~\citep{llava-ov}, and Qwen-2.5-VL-7B~\citep{qwen2_5vl}) and recent reinforcement learning with verifiable reward (RLVR) approaches (e.g., R1-VL~\citep{r1_vl}, Vision-R1~\citep{vision_r1}, R1-OneVision~\citep{r1_onevision}, OpenVLThinker~\citep{openvlthinker}, MM-Eureka~\citep{Mm-eureka}, ThinkLite~\citep{thinklite}, VLAA-Thinker~\citep{vlaa_thinking}, and NoisyRollout~\citep{noisyrollout}). Results are summarized in Table~\ref{tab:main_result}.

$\bullet$ \textbf{Comparison with RLVR baselines.}
Under the same training data scale (2.1K Geo3K samples), ToR consistently improves strong RLVR baselines.
Applying token re-weighting to GRPO improves MathVerse from 50.8 to 53.0 and HalluBench from 69.8 to 72.4.
Similarly, ToR-DAPO outperforms DAPO across all benchmarks, with particularly notable gains on WeMath and MathVista.

$\bullet$ \textbf{Generalization across data and model scales.}
ToR continues to provide consistent improvements when scaling the training data from Geo3K to ViRL-39K, achieving the strongest overall performance on Qwen-2.5-VL-7B across all benchmarks.
Moreover, ToR generalizes to a smaller Qwen-2.5-VL-3B backbone, consistently improving over the DAPO baseline despite reduced model capacity.

Overall, Table~\ref{tab:main_result} shows that token re-weighting demonstrates robust and generalizable improvements to RLVR-based multimodal learning, yielding consistent gains across model sizes, training data scales, and diverse reasoning and perception benchmarks.
Importantly, all ToR results are obtained using the same hyperparameter settings across different model backbones and data scales.
\section{Related Work}
\label{sec:related_work}

In this section, we first briefly summarize recent advances in computer vision~\citep{Adavip, lu_mm, dama, lu2024rethinking}, especially those that focus on enhancing reasoning capabilities in Multi-modal LLMs, and then, we illustrate related methods for reinforcement learning with verifiable rewards. Finally, we enumerate the differences between our approach and related methods.

\subsection{Reasoning in Multi-modal LLMs}
Existing approaches that focus on reasoning in multi-modal LLMs can be broadly categorized into:

$\bullet$ \textbf{Extending reasoning LLMs with visual understanding.}
Building upon the strong reasoning capabilities of recent LLMs, one research branch explores incorporating visual content into LLMs. 
Typical strategies include:
\textbf{(1)} integrating visual encoders into LLMs to directly extend them to multimodal scenarios~\citep{skywork_r1v, skywork_r1v2}; 
and \textbf{(2)} transforming images into captions and feeding them into LLMs to bridge perception and reasoning~\citep{vision_r1, PAPO}.

$\bullet$ \textbf{Enhancing reasoning abilities within MLLMs.}
Another research branch seeks to endow existing MLLMs with reasoning skills. Representative approaches include: 
\textbf{(1)} transferring reasoning priors from reasoning LLMs into MLLMs through model merging, thereby leveraging the complementary strengths of both models~\citep{bring_reason_to_vision}; 
and \textbf{(2)} adapting RLVR algorithms, such as Group Relative Policy Optimization (GRPO)~\citep{deepseekmath}, to enhance reasoning capabilities in multimodal settings~\citep{vision_r1, visual_rft, durian_Li}.

\begin{figure*}[t]
\centering
\includegraphics[width=\textwidth]{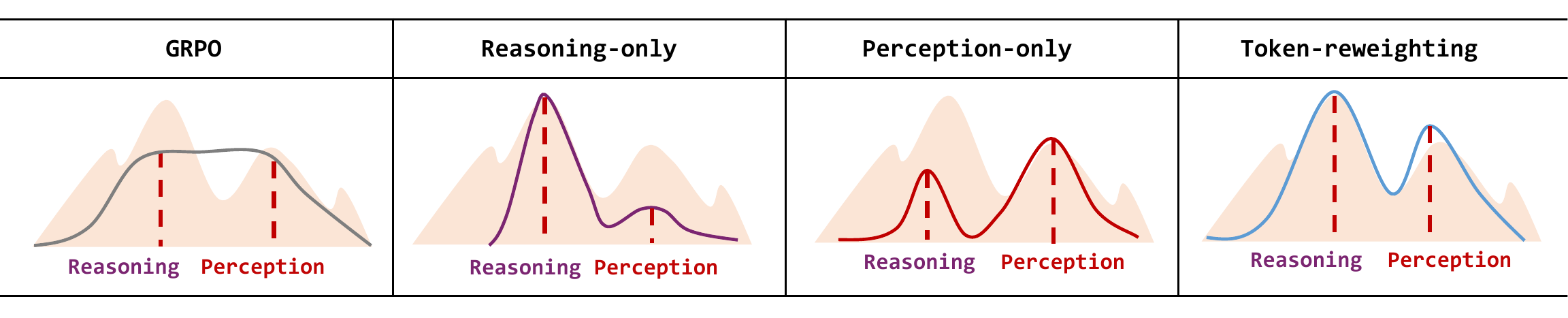}
\caption{\textbf{Comparison of optimization behaviors under different token selection strategies.}
Vanilla GRPO optimizes uniformly across all tokens.
Reasoning-only and perception-only optimization concentrate on a single token type, leading to imbalanced training.
Token reweighting jointly emphasizes reasoning- and perception-related tokens, achieving a more balanced optimization.}
\label{fig:analysis}
\end{figure*}

\subsection{Reinforcement Learning with Verifiable Rewards}
RLVR has recently demonstrated its effectiveness in aligning LLMs with verifiable reasoning outcomes. 
Among the implementations, GRPO~\citep{deepseekmath} has shown great success with more stable advantage estimation. Subsequent refinements have further improved its efficiency and effectiveness~\citep{dapo, dr_grpo, 80_20_tokens}.
When extending RLVR to multimodal settings, current methods primarily focus on two distinct branches:

$\bullet$ \textbf{Emphasizing the importance of visual understanding within RLVR.} Specifically, works like~\citep{perception_r1, perception_r1_2} introduce perception-oriented rewards to incentivize accurate visual grounding and understanding. Moreover, works such as~\citep{noisyrollout, PAPO, vision_matters} employ data augmentation strategies to improve model robustness and sensitivity against visual variations. Works like~\citep{thyme, visual_arft} incorporate image manipulation tools like cropping and zooming in to focus on critical regions in the image during the reasoning process.

$\bullet$ \textbf{Constructing coherent reasoning chains with RLVR.} Representative works include: \textbf{(1)} distilling chain-of-thought reasoning patterns from stronger reasoning models to improve reasoning coherence in multimodal tasks~\citep{vision_r1, cold_start}; and \textbf{(2)} modifying different components of GRPO (e.g., clip ratios or advantage estimation) to emphasize reasoning-critical tokens better and stabilize training~\citep{r1_reward, Mm-eureka, vl_rethinker}.

\noindent \textbf{Differences.}
Unlike existing methods that focus on optimizing either perception or reasoning abilities in isolation, our work systematically investigates and addresses the interdependence between these two capabilities. 
Specifically, through comprehensive token-level analysis, we demonstrate that perception and reasoning tokens exhibit complex interactions during training, where optimizing one type can inadvertently impair the other. 
To address this challenge, we propose a simple token-reweighting strategy that explicitly balances the optimization of both perception and reasoning tokens, leading to significant performance improvements across both capabilities.
\section{Discussion}
\textbf{Understanding Token Reweighting.} Unlike LLMs, MLLMs generate responses in which perception- and reasoning-related tokens are naturally interleaved.
These tokens play different roles: perception-related tokens are responsible for grounding the visual inputs, while reasoning-related tokens determine key reasoning decisions.
As illustrated in Figure~\ref{fig:analysis}, optimizing all tokens uniformly, as in vanilla GRPO, ignores this distinction and spreads learning signals over tokens that contribute little to reasoning or perception improvement.

Optimizing only reasoning-related tokens focuses on uncertain decision points and can improve reasoning quality, but it neglects perceptual tokens and weakens visual grounding.
In contrast, optimizing only perception-related tokens strengthens visual grounding but leaves reasoning uncertainty largely unchanged, limiting reasoning capacities.

Token reweighting balances these two extremes by jointly emphasizing both reasoning- and perception-related tokens.
This allows the model to improve critical reasoning decisions while still maintaining strong visual grounding, resulting in more stable and effective training across various backbones and data scales in practice.

\section{Conclusion}
In this work, through systematic token-wise analysis, we uncover a fundamental challenge in extending RLVR to multimodal LLMs: \textbf{the intrinsic interdependence between perception and reasoning.} We show that overemphasizing either capability inevitably impairs the other, yet current approaches overlook this issue and optimize them in isolation.

To address this, we proposed a simple yet effective \textbf{To}ken-\textbf{R}eweighting (\textbf{ToR}) strategy that identifies and reweights perception- and reasoning-related tokens during RLVR training. We apply ToR over current RLVR algorithms (\textit{e.g.}, GRPO and DAPO), and achieve significant performance gains across diverse benchmarks, consistently enhancing both perception and reasoning.

\noindent \textbf{Limitations and Future Work.}
While ToR establishes a foundation for addressing perception-reasoning interdependence, several promising directions remain open:
\ding{182} \textbf{Fine-grained token identification strategies}: Precisely localizing critical regions in images using models like SAM~\citep{sam} to identify more fine-grained perception-critical tokens.
\ding{183} \textbf{Dynamic token reweighting}: Dynamically assigning weights to tokens based on their gradient contributions or connections to final outcomes~\citep{rlpr}.
\ding{184} \textbf{Extending beyond tokens}: As tokens derive meaning from context, future work could explore perception-reasoning interdependence at broader contextual levels --- optimizing tokens within their semantic context, preserving their contextual relationships.
\ding{185} \textbf{Exploring broader applications}: Extending this interdependence framework to more complex scenarios, \textit{e.g.}, unified multi-modal generation and understanding tasks~\citep{qwen_image, bagel}, where visual tokens participate in reasoning processes.

\newpage
\section*{Impact Statements}
This paper presents work whose goal is to advance the field of machine learning. There are many potential societal consequences of our work, none of which we feel must be specifically highlighted here.

\bibliography{main}

@inproceedings{tulu3,
title={Tulu 3: Pushing Frontiers in Open Language Model Post-Training},
author={Nathan Lambert and Jacob Morrison and Valentina Pyatkin and Shengyi Huang and Hamish Ivison and Faeze Brahman and Lester James Validad Miranda and Alisa Liu and Nouha Dziri and Xinxi Lyu and Yuling Gu and Saumya Malik and Victoria Graf and Jena D. Hwang and Jiangjiang Yang and Ronan Le Bras and Oyvind Tafjord and Christopher Wilhelm and Luca Soldaini and Noah A. Smith and Yizhong Wang and Pradeep Dasigi and Hannaneh Hajishirzi},
booktitle={Second Conference on Language Modeling},
year={2025},
url={https://openreview.net/forum?id=i1uGbfHHpH}
}

@article{deepseek_r1,
  title={Deepseek-r1 incentivizes reasoning in llms through reinforcement learning},
  author={Guo, Daya and Yang, Dejian and Zhang, Haowei and Song, Junxiao and Wang, Peiyi and Zhu, Qihao and Xu, Runxin and Zhang, Ruoyu and Ma, Shirong and Bi, Xiao and others},
  journal={Nature},
  volume={645},
  number={8081},
  pages={633--638},
  year={2025},
  publisher={Nature Publishing Group UK London}
}

@InProceedings{visual_rft,
    author    = {Liu, Ziyu and Sun, Zeyi and Zang, Yuhang and Dong, Xiaoyi and Cao, Yuhang and Duan, Haodong and Lin, Dahua and Wang, Jiaqi},
    title     = {Visual-RFT: Visual Reinforcement Fine-Tuning},
    booktitle = {Proceedings of the IEEE/CVF International Conference on Computer Vision (ICCV)},
    month     = {October},
    year      = {2025},
    pages     = {2034-2044}
}

@article{r1_reward,
  title={R1-reward: Training multimodal reward model through stable reinforcement learning},
  author={Zhang, Yi-Fan and Lu, Xingyu and Hu, Xiao and Fu, Chaoyou and Wen, Bin and Zhang, Tianke and Liu, Changyi and Jiang, Kaiyu and Chen, Kaibing and Tang, Kaiyu and others},
  journal={arXiv preprint arXiv:2505.02835},
  year={2025}
}

@article{vision_r1,
  title={Vision-r1: Incentivizing reasoning capability in multimodal large language models},
  author={Huang, Wenxuan and Jia, Bohan and Zhai, Zijie and Cao, Shaosheng and Ye, Zheyu and Zhao, Fei and Xu, Zhe and Hu, Yao and Lin, Shaohui},
  journal={arXiv preprint arXiv:2503.06749},
  year={2025}
}

@article{deepseekmath,
  title={Deepseekmath: Pushing the limits of mathematical reasoning in open language models},
  author={Shao, Zhihong and Wang, Peiyi and Zhu, Qihao and Xu, Runxin and Song, Junxiao and Bi, Xiao and Zhang, Haowei and Zhang, Mingchuan and Li, YK and Wu, Yang and others},
  journal={arXiv preprint arXiv:2402.03300},
  year={2024}
}

@article{cold_start,
  title={Advancing Multimodal Reasoning via Reinforcement Learning with Cold Start},
  author={Wei, Lai and Li, Yuting and Zheng, Kaipeng and Wang, Chen and Wang, Yue and Kong, Linghe and Sun, Lichao and Huang, Weiran},
  journal={arXiv preprint arXiv:2505.22334},
  year={2025}
}

@article{Mm-eureka,
  title={Mm-eureka: Exploring visual aha moment with rule-based large-scale reinforcement learning},
  author={Meng, Fanqing and Du, Lingxiao and Liu, Zongkai and Zhou, Zhixiang and Lu, Quanfeng and Fu, Daocheng and Shi, Botian and Wang, Wenhai and He, Junjun and Zhang, Kaipeng and others},
  journal={CoRR},
  year={2025}
}

@article{PAPO,
  title={Perception-aware policy optimization for multimodal reasoning},
  author={Wang, Zhenhailong and Guo, Xuehang and Stoica, Sofia and Xu, Haiyang and Wang, Hongru and Ha, Hyeonjeong and Chen, Xiusi and Chen, Yangyi and Yan, Ming and Huang, Fei and others},
  journal={arXiv preprint arXiv:2507.06448},
  year={2025}
}

@inproceedings{noisyrollout,
title={NoisyRollout: Reinforcing Visual Reasoning with Data Augmentation},
author={Xiangyan Liu and Jinjie Ni and Zijian Wu and Chao Du and Longxu Dou and Haonan Wang and Tianyu Pang and Michael Qizhe Shieh},
booktitle={2nd AI for Math Workshop @ ICML 2025},
year={2025},
url={https://openreview.net/forum?id=GdNm9PEAei}
}

@inproceedings{80_20_tokens,
title={Beyond the 80/20 Rule: High-Entropy Minority Tokens Drive Effective Reinforcement Learning for {LLM} Reasoning},
author={Shenzhi Wang and Le Yu and Chang Gao and Chujie Zheng and Shixuan Liu and Rui Lu and Kai Dang and Xiong-Hui Chen and Jianxin Yang and Zhenru Zhang and Yuqiong Liu and An Yang and Andrew Zhao and Yang Yue and Shiji Song and Bowen Yu and Gao Huang and Junyang Lin},
booktitle={The Thirty-ninth Annual Conference on Neural Information Processing Systems},
year={2025},
url={https://openreview.net/forum?id=yfcpdY4gMP}
}

@article{llm_reasoning_with_exploration,
  title={Reasoning with exploration: An entropy perspective},
  author={Cheng, Daixuan and Huang, Shaohan and Zhu, Xuekai and Dai, Bo and Zhao, Wayne Xin and Zhang, Zhenliang and Wei, Furu},
  journal={arXiv preprint arXiv:2506.14758},
  year={2025}
}

@article{qwen3,
  title={Qwen3 technical report},
  author={Yang, An and Li, Anfeng and Yang, Baosong and Zhang, Beichen and Hui, Binyuan and Zheng, Bo and Yu, Bowen and Gao, Chang and Huang, Chengen and Lv, Chenxu and others},
  journal={arXiv preprint arXiv:2505.09388},
  year={2025}
}

@article{dapo,
  title={Dapo: An open-source llm reinforcement learning system at scale},
  author={Yu, Qiying and Zhang, Zheng and Zhu, Ruofei and Yuan, Yufeng and Zuo, Xiaochen and Yue, Yu and Dai, Weinan and Fan, Tiantian and Liu, Gaohong and Liu, Lingjun and others},
  journal={arXiv preprint arXiv:2503.14476},
  year={2025}
}

@article{skywork_r1v,
  title={Skywork r1v: Pioneering multimodal reasoning with chain-of-thought},
  author={Peng, Yi and Wang, Peiyu and Wang, Xiaokun and Wei, Yichen and Pei, Jiangbo and Qiu, Weijie and Jian, Ai and Hao, Yunzhuo and Pan, Jiachun and Xie, Tianyidan and others},
  journal={arXiv preprint arXiv:2504.05599},
  year={2025}
}

@article{skywork_r1v2,
  title={Skywork r1v2: Multimodal hybrid reinforcement learning for reasoning},
  author={Wang, Peiyu and Wei, Yichen and Peng, Yi and Wang, Xiaokun and Qiu, Weijie and Shen, Wei and Xie, Tianyidan and Pei, Jiangbo and Zhang, Jianhao and Hao, Yunzhuo and others},
  journal={arXiv preprint arXiv:2504.16656},
  year={2025}
}

@inproceedings{bring_reason_to_vision,
title={Bring Reason to Vision: Understanding Perception and Reasoning through Model Merging},
author={Shiqi Chen and Jinghan Zhang and Tongyao Zhu and Wei Liu and Siyang Gao and Miao Xiong and Manling Li and Junxian He},
booktitle={Forty-second International Conference on Machine Learning},
year={2025},
url={https://openreview.net/forum?id=ntCAP6tMoX}
}

@inproceedings{dr_grpo,
title={Understanding R1-Zero-Like Training: A Critical Perspective},
author={Zichen Liu and Changyu Chen and Wenjun Li and Penghui Qi and Tianyu Pang and Chao Du and Wee Sun Lee and Min Lin},
booktitle={2nd AI for Math Workshop @ ICML 2025},
year={2025},
url={https://openreview.net/forum?id=jLpC1zavzn}
}

@article{perception_r1,
  title={Advancing Multimodal Reasoning Capabilities of Multimodal Large Language Models via Visual Perception Reward},
  author={Xiao, Tong and Xu, Xin and Huang, Zhenya and Gao, Hongyu and Liu, Quan and Liu, Qi and Chen, Enhong},
  journal={arXiv preprint arXiv:2506.07218},
  year={2025}
}

@inproceedings{perception_r1_2,
title={Perception-R1: Pioneering Perception Policy with Reinforcement Learning},
author={En Yu and Kangheng Lin and Liang Zhao and jisheng yin and Yana Wei and Yuang Peng and Haoran Wei and Jianjian Sun and Chunrui Han and Zheng Ge and Xiangyu Zhang and Daxin Jiang and Jingyu Wang and Wenbing Tao},
booktitle={The Thirty-ninth Annual Conference on Neural Information Processing Systems},
year={2025},
url={https://openreview.net/forum?id=BeXcXrXetA}
}

@article{vision_matters,
  title={Vision Matters: Simple Visual Perturbations Can Boost Multimodal Math Reasoning},
  author={Li, Yuting and Wei, Lai and Zheng, Kaipeng and Huang, Jingyuan and Kong, Linghe and Sun, Lichao and Huang, Weiran},
  journal={arXiv preprint arXiv:2506.09736},
  year={2025}
}

@article{thyme,
  title={Thyme: Think Beyond Images},
  author={Zhang, Yi-Fan and Lu, Xingyu and Yin, Shukang and Fu, Chaoyou and Chen, Wei and Hu, Xiao and Wen, Bin and Jiang, Kaiyu and Liu, Changyi and Zhang, Tianke and others},
  journal={arXiv preprint arXiv:2508.11630},
  year={2025}
}

@article{visual_arft,
  title={Visual Agentic Reinforcement Fine-Tuning},
  author={Liu, Ziyu and Zang, Yuhang and Zou, Yushan and Liang, Zijian and Dong, Xiaoyi and Cao, Yuhang and Duan, Haodong and Lin, Dahua and Wang, Jiaqi},
  journal={arXiv preprint arXiv:2505.14246},
  year={2025}
}

@inproceedings{vl_rethinker,
title={{VL}-Rethinker: Incentivizing Self-Reflection of Vision-Language Models with Reinforcement Learning},
author={Haozhe Wang and Chao Qu and Zuming Huang and Wei Chu and Fangzhen Lin and Wenhu Chen},
booktitle={The Thirty-ninth Annual Conference on Neural Information Processing Systems},
year={2025},
url={https://openreview.net/forum?id=4oYxzssbVg}
}

@inproceedings{sam,
  title={Segment anything},
  author={Kirillov, Alexander and Mintun, Eric and Ravi, Nikhila and Mao, Hanzi and Rolland, Chloe and Gustafson, Laura and Xiao, Tete and Whitehead, Spencer and Berg, Alexander C and Lo, Wan-Yen and others},
  booktitle={Proceedings of the IEEE/CVF international conference on computer vision},
  pages={4015--4026},
  year={2023}
}

@article{rlpr,
  title={RLPR: Extrapolating RLVR to General Domains without Verifiers},
  author={Yu, Tianyu and Ji, Bo and Wang, Shouli and Yao, Shu and Wang, Zefan and Cui, Ganqu and Yuan, Lifan and Ding, Ning and Yao, Yuan and Liu, Zhiyuan and others},
  journal={arXiv preprint arXiv:2506.18254},
  year={2025}
}

@article{qwen_image,
  title={Qwen-image technical report},
  author={Wu, Chenfei and Li, Jiahao and Zhou, Jingren and Lin, Junyang and Gao, Kaiyuan and Yan, Kun and Yin, Sheng-ming and Bai, Shuai and Xu, Xiao and Chen, Yilei and others},
  journal={arXiv preprint arXiv:2508.02324},
  year={2025}
}

@article{bagel,
  title={Emerging properties in unified multimodal pretraining},
  author={Deng, Chaorui and Zhu, Deyao and Li, Kunchang and Gou, Chenhui and Li, Feng and Wang, Zeyu and Zhong, Shu and Yu, Weihao and Nie, Xiaonan and Song, Ziang and others},
  journal={arXiv preprint arXiv:2505.14683},
  year={2025}
}

@article{wemath,
  title={We-math: Does your large multimodal model achieve human-like mathematical reasoning?},
  author={Qiao, Runqi and Tan, Qiuna and Dong, Guanting and Wu, Minhui and Sun, Chong and Song, Xiaoshuai and GongQue, Zhuoma and Lei, Shanglin and Wei, Zhe and Zhang, Miaoxuan and others},
  journal={arXiv preprint arXiv:2407.01284},
  year={2024}
}

@inproceedings{mathverse,
  title={Mathverse: Does your multi-modal llm truly see the diagrams in visual math problems?},
  author={Zhang, Renrui and Jiang, Dongzhi and Zhang, Yichi and Lin, Haokun and Guo, Ziyu and Qiu, Pengshuo and Zhou, Aojun and Lu, Pan and Chang, Kai-Wei and Qiao, Yu and others},
  booktitle={European Conference on Computer Vision},
  pages={169--186},
  year={2024},
  organization={Springer}
}

@article{mathvision,
  title={Measuring multimodal mathematical reasoning with math-vision dataset},
  author={Wang, Ke and Pan, Junting and Shi, Weikang and Lu, Zimu and Ren, Houxing and Zhou, Aojun and Zhan, Mingjie and Li, Hongsheng},
  journal={Advances in Neural Information Processing Systems},
  volume={37},
  pages={95095--95169},
  year={2024}
}

@inproceedings{mathvista,
title={MathVista: Evaluating Mathematical Reasoning of Foundation Models in Visual Contexts},
author={Pan Lu and Hritik Bansal and Tony Xia and Jiacheng Liu and Chunyuan Li and Hannaneh Hajishirzi and Hao Cheng and Kai-Wei Chang and Michel Galley and Jianfeng Gao},
booktitle={The Twelfth International Conference on Learning Representations},
year={2024},
url={https://openreview.net/forum?id=KUNzEQMWU7}
}

@inproceedings{hallusionbench,
  title={Hallusionbench: an advanced diagnostic suite for entangled language hallucination and visual illusion in large vision-language models},
  author={Guan, Tianrui and Liu, Fuxiao and Wu, Xiyang and Xian, Ruiqi and Li, Zongxia and Liu, Xiaoyu and Wang, Xijun and Chen, Lichang and Huang, Furong and Yacoob, Yaser and others},
  booktitle={Proceedings of the IEEE/CVF Conference on Computer Vision and Pattern Recognition},
  pages={14375--14385},
  year={2024}
}

@inproceedings{geometry3k,
  title={Inter-GPS: Interpretable Geometry Problem Solving with Formal Language and Symbolic Reasoning},
  author={Lu, Pan and Gong, Ran and Jiang, Shibiao and Qiu, Liang and Huang, Siyuan and Liang, Xiaodan and Zhu, Song-chun},
  booktitle={Proceedings of the 59th Annual Meeting of the Association for Computational Linguistics and the 11th International Joint Conference on Natural Language Processing (Volume 1: Long Papers)},
  pages={6774--6786},
  year={2021}
}

@misc{easyr1,
  title        = {EasyR1: An Efficient, Scalable, Multi-Modality RL Training Framework},
  author       = {Yaowei, Zheng and Junting, Lu and Shenzhi, Wang and Zhangchi, Feng and Dongdong, Kuang and Yuwen, Xiong},
  howpublished = {\url{https://github.com/hiyouga/EasyR1}},
  year         = {2025}
}

@article{qwen2_5vl,
  title={Qwen2. 5-vl technical report},
  author={Bai, Shuai and Chen, Keqin and Liu, Xuejing and Wang, Jialin and Ge, Wenbin and Song, Sibo and Dang, Kai and Wang, Peng and Wang, Shijie and Tang, Jun and others},
  journal={arXiv preprint arXiv:2502.13923},
  year={2025}
}

@article{internvl2.5,
  title={Expanding performance boundaries of open-source multimodal models with model, data, and test-time scaling},
  author={Chen, Zhe and Wang, Weiyun and Cao, Yue and Liu, Yangzhou and Gao, Zhangwei and Cui, Erfei and Zhu, Jinguo and Ye, Shenglong and Tian, Hao and Liu, Zhaoyang and others},
  journal={arXiv preprint arXiv:2412.05271},
  year={2024}
}

@article{llava-ov,
title={{LL}a{VA}-OneVision: Easy Visual Task Transfer},
author={Bo Li and Yuanhan Zhang and Dong Guo and Renrui Zhang and Feng Li and Hao Zhang and Kaichen Zhang and Peiyuan Zhang and Yanwei Li and Ziwei Liu and Chunyuan Li},
journal={Transactions on Machine Learning Research},
issn={2835-8856},
year={2025},
url={https://openreview.net/forum?id=zKv8qULV6n},
note={}
}

@article{internvl3,
  title={Internvl3: Exploring advanced training and test-time recipes for open-source multimodal models},
  author={Zhu, Jinguo and Wang, Weiyun and Chen, Zhe and Liu, Zhaoyang and Ye, Shenglong and Gu, Lixin and Tian, Hao and Duan, Yuchen and Su, Weijie and Shao, Jie and others},
  journal={arXiv preprint arXiv:2504.10479},
  year={2025}
}

@InProceedings{r1_vl,
    author    = {Zhang, Jingyi and Huang, Jiaxing and Yao, Huanjin and Liu, Shunyu and Zhang, Xikun and Lu, Shijian and Tao, Dacheng},
    title     = {R1-VL: Learning to Reason with Multimodal Large Language Models via Step-wise Group Relative Policy Optimization},
    booktitle = {Proceedings of the IEEE/CVF International Conference on Computer Vision (ICCV)},
    month     = {October},
    year      = {2025},
    pages     = {1859-1869}
}

@InProceedings{r1_onevision,
    author    = {Yang, Yi and He, Xiaoxuan and Pan, Hongkun and Jiang, Xiyan and Deng, Yan and Yang, Xingtao and Lu, Haoyu and Yin, Dacheng and Rao, Fengyun and Zhu, Minfeng and Zhang, Bo and Chen, Wei},
    title     = {R1-Onevision: Advancing Generalized Multimodal Reasoning through Cross-Modal Formalization},
    booktitle = {Proceedings of the IEEE/CVF International Conference on Computer Vision (ICCV)},
    month     = {October},
    year      = {2025},
    pages     = {2376-2385}
}

@inproceedings{openvlthinker,
title={Open{VLT}hinker: Complex Vision-Language Reasoning via Iterative {SFT}-{RL} Cycles},
author={Yihe Deng and Hritik Bansal and Fan Yin and Nanyun Peng and Wei Wang and Kai-Wei Chang},
booktitle={The Thirty-ninth Annual Conference on Neural Information Processing Systems},
year={2025},
url={https://openreview.net/forum?id=gfX1nqBKtu}
}

@inproceedings{thinklite,
title={So{TA} with Less: {MCTS}-Guided Sample Selection for Data-Efficient Visual Reasoning Self-Improvement},
author={Xiyao Wang and Zhengyuan Yang and Chao Feng and Hongjin Lu and Linjie Li and Chung-Ching Lin and Kevin Lin and Furong Huang and Lijuan Wang},
booktitle={The Thirty-ninth Annual Conference on Neural Information Processing Systems},
year={2025},
url={https://openreview.net/forum?id=PHu9xJeAum}
}

@article{vlaa_thinking,
title={{SFT} or {RL}? An Early Investigation into Training R1-Like Reasoning Large Vision-Language Models},
author={Hardy Chen and Haoqin Tu and Fali Wang and Hui Liu and Xianfeng Tang and Xinya Du and Yuyin Zhou and Cihang Xie},
journal={Transactions on Machine Learning Research},
issn={2835-8856},
year={2025},
url={https://openreview.net/forum?id=wZI5qkQeDF},
note={}
}

@misc{durian_Li,
      title={Enhancing Multi-Modal LLMs Reasoning via Difficulty-Aware Group Normalization}, 
      author={Jinghan Li and Junfeng Fang and Jinda Lu and Yuan Wang and Xiaoyan Guo and Tianyu Zhang and Xiang Wang and Xiangnan He},
      year={2026},
      eprint={2602.21743},
      archivePrefix={arXiv},
      primaryClass={cs.CV},
      url={https://arxiv.org/abs/2602.21743}, 
}

@misc{Adavip,
      title={AdaViP: Aligning Multi-modal LLMs via Adaptive Vision-enhanced Preference Optimization}, 
      author={Jinda Lu and Jinghan Li and Yuan Gao and Junkang Wu and Jiancan Wu and Xiang Wang and Xiangnan He},
      year={2025},
      eprint={2504.15619},
      archivePrefix={arXiv},
      primaryClass={cs.CV},
      url={https://arxiv.org/abs/2504.15619}, 
}

@inproceedings{lu_mm,
  title={Semantic-based selection, synthesis, and supervision for few-shot learning},
  author={Lu, Jinda and Wang, Shuo and Zhang, Xinyu and Hao, Yanbin and He, Xiangnan},
  booktitle={Proceedings of the 31st ACM International Conference on Multimedia},
  pages={3569--3578},
  year={2023}
}

@inproceedings{dama,
  title={DAMA: Data-and Model-aware Alignment of Multi-modal LLMs},
  author={Lu, Jinda and Wu, Junkang and Li, Jinghan and Jia, Xiaojun and Wang, Shuo and Zhang, Yifan and Fang, Junfeng and Wang, Xiang and He, Xiangnan},
  booktitle={International Conference on Machine Learning},
  pages={40726--40740},
  year={2025},
  organization={PMLR}
}

@article{lu2024rethinking,
  title={Rethinking visual content refinement in low-shot clip adaptation},
  author={Lu, Jinda and Wang, Shuo and Hao, Yanbin and Liu, Haifeng and Wang, Xiang and Wang, Meng},
  journal={arXiv preprint arXiv:2407.14117},
  year={2024}
}
\bibliographystyle{icml2026}

\newpage
\appendix
\onecolumn
\section{Token Distribution and Characteristic Analysis}

\begin{figure}[htbp]
    \centering
    \begin{minipage}{0.45\textwidth}
        \centering
        \includegraphics[width=\textwidth]{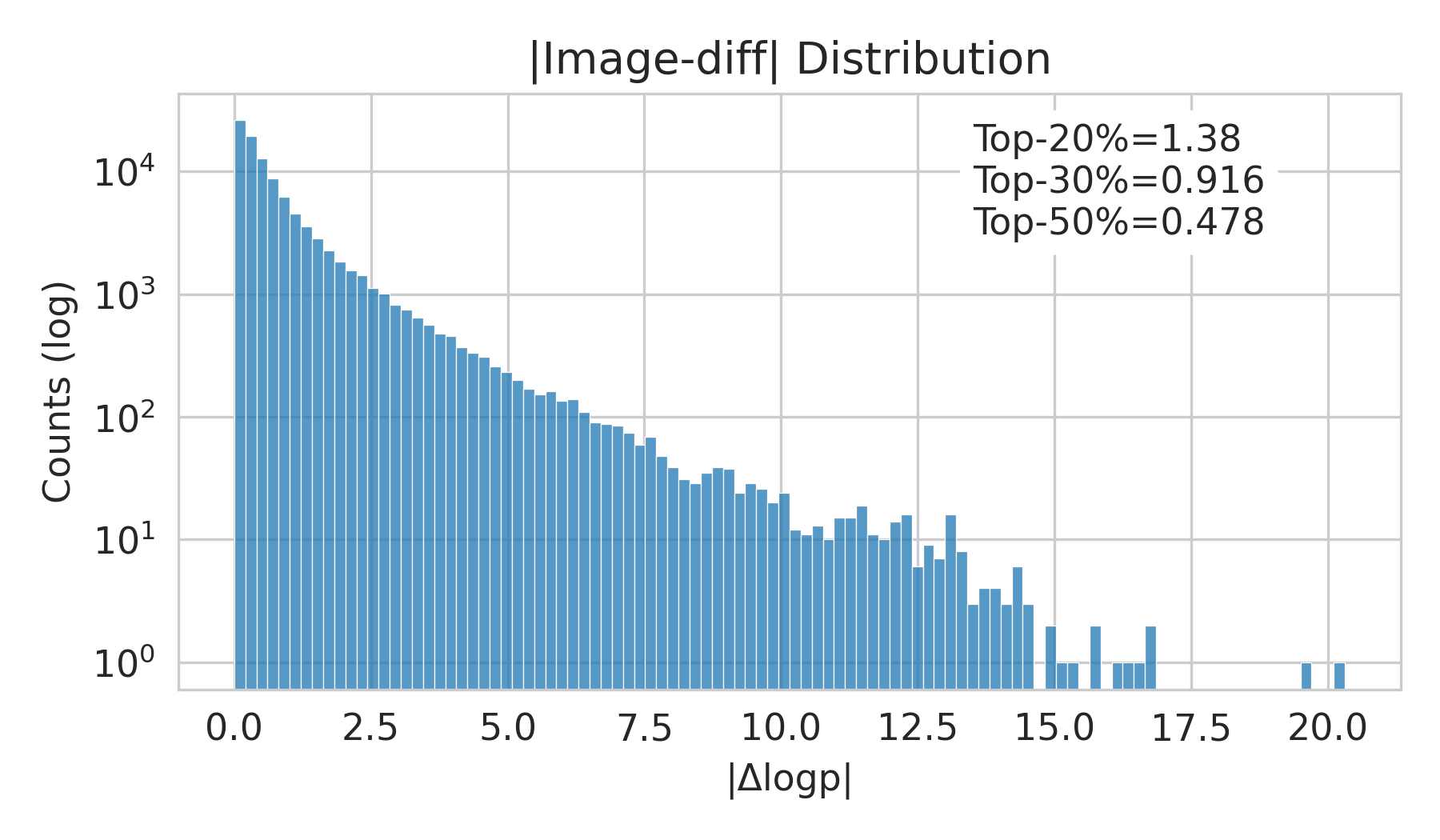}
        \caption{Distribution of log-probability differences for Qwen-2.5-VL-7B on HallusionBench, used to identify perception-related tokens~\citep{hallusionbench}.}
        \label{fig:logp-diff-hallubench}
    \end{minipage}
    \hfill
    \begin{minipage}{0.45\textwidth}
        \centering
        \includegraphics[width=\textwidth]{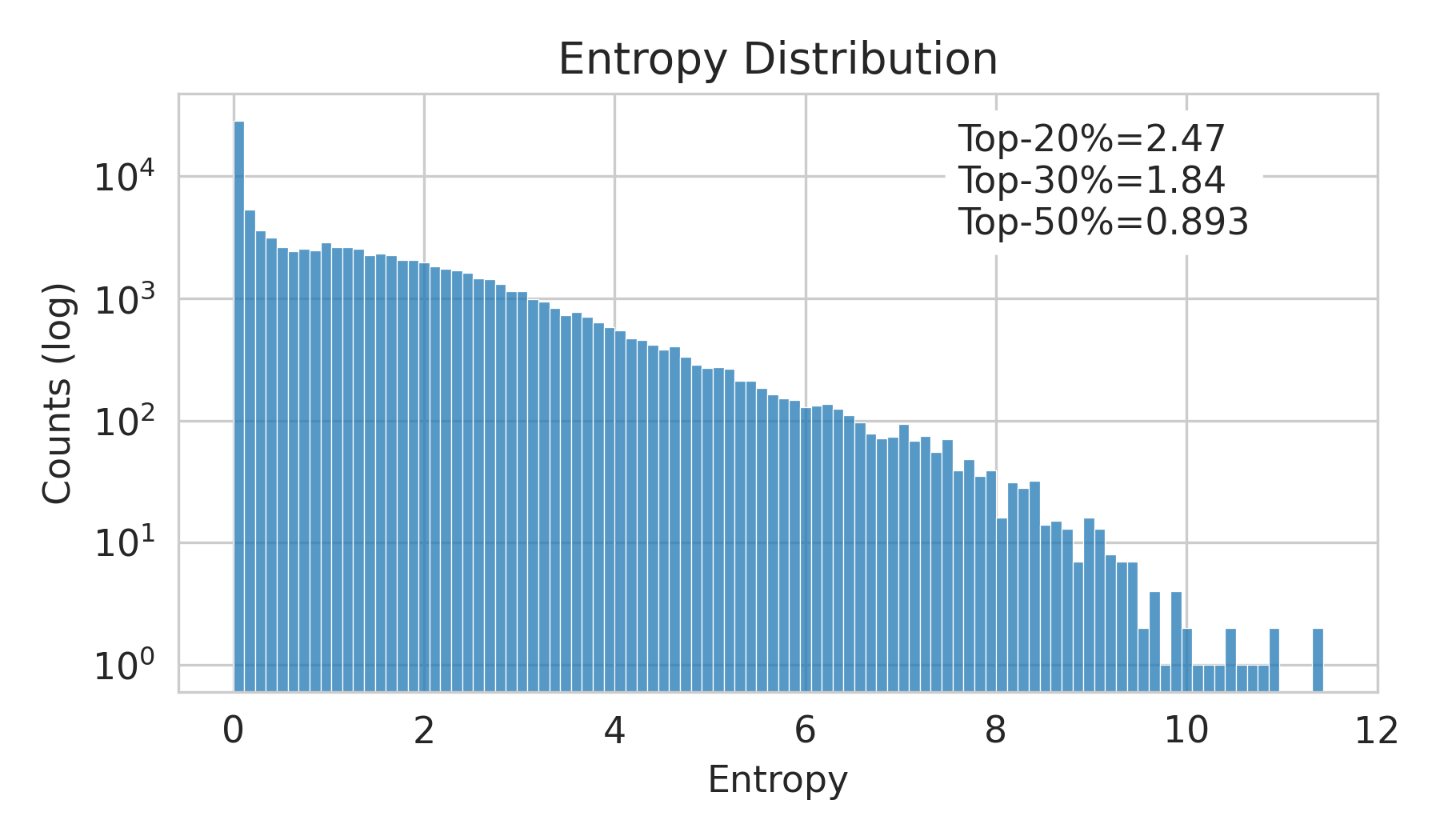}
        \caption{Distribution of entropy values for Qwen-2.5-VL-7B on HallusionBench, used to identify reasoning-related tokens~\citep{hallusionbench}.}
        \label{fig:entropy-hallubench}
    \end{minipage}
\end{figure}

\begin{figure}[htbp]
    \centering
    \begin{minipage}{0.45\textwidth}
        \centering
        \includegraphics[width=\textwidth]{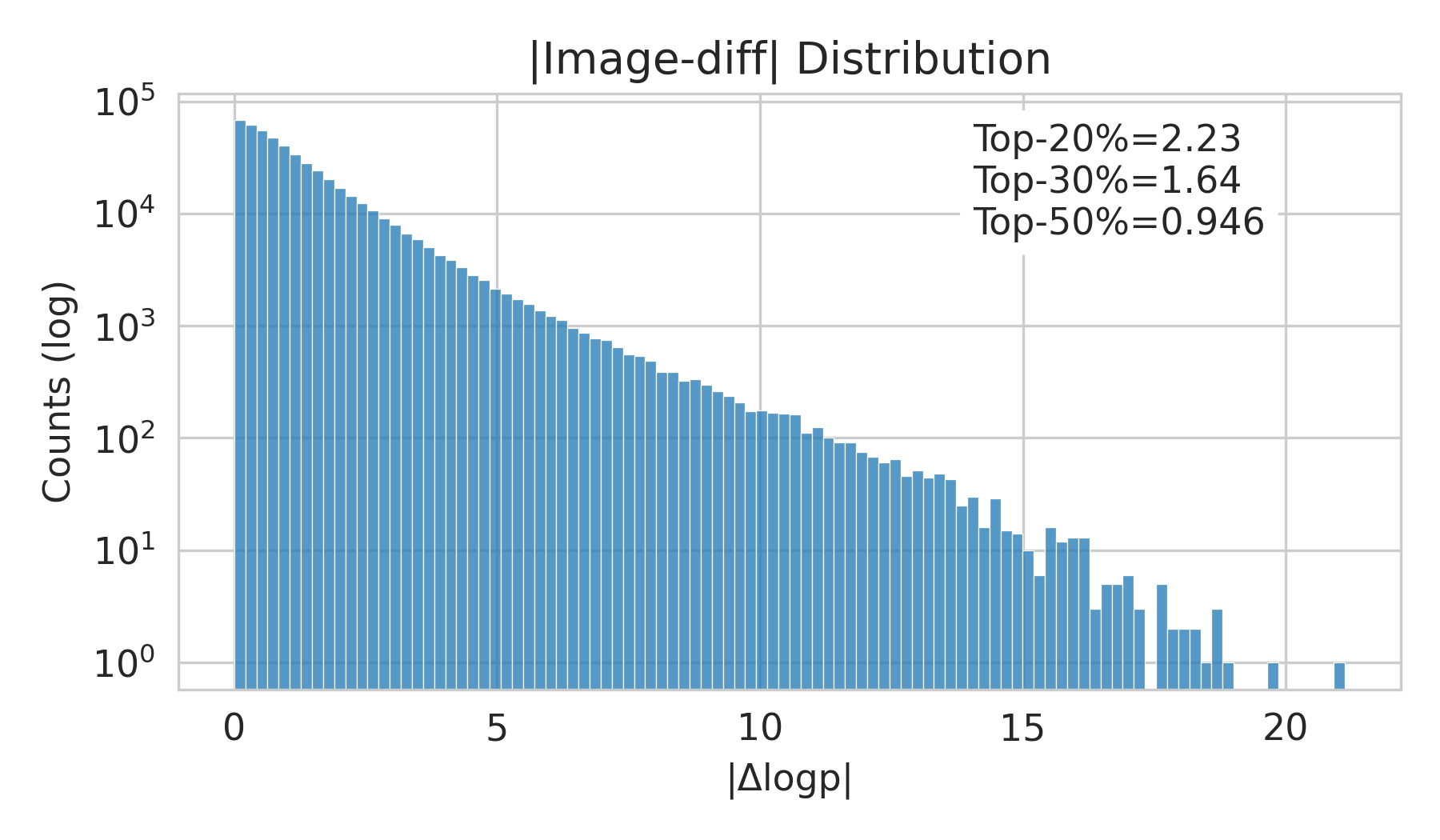}
        \caption{Distribution of log-probability differences for Qwen-2.5-VL-7B on WeMath, used to identify perception-related tokens~\citep{wemath}.}
        \label{fig:logp-diff-wemath}
    \end{minipage}
    \hfill
    \begin{minipage}{0.45\textwidth}
        \centering
        \includegraphics[width=\textwidth]{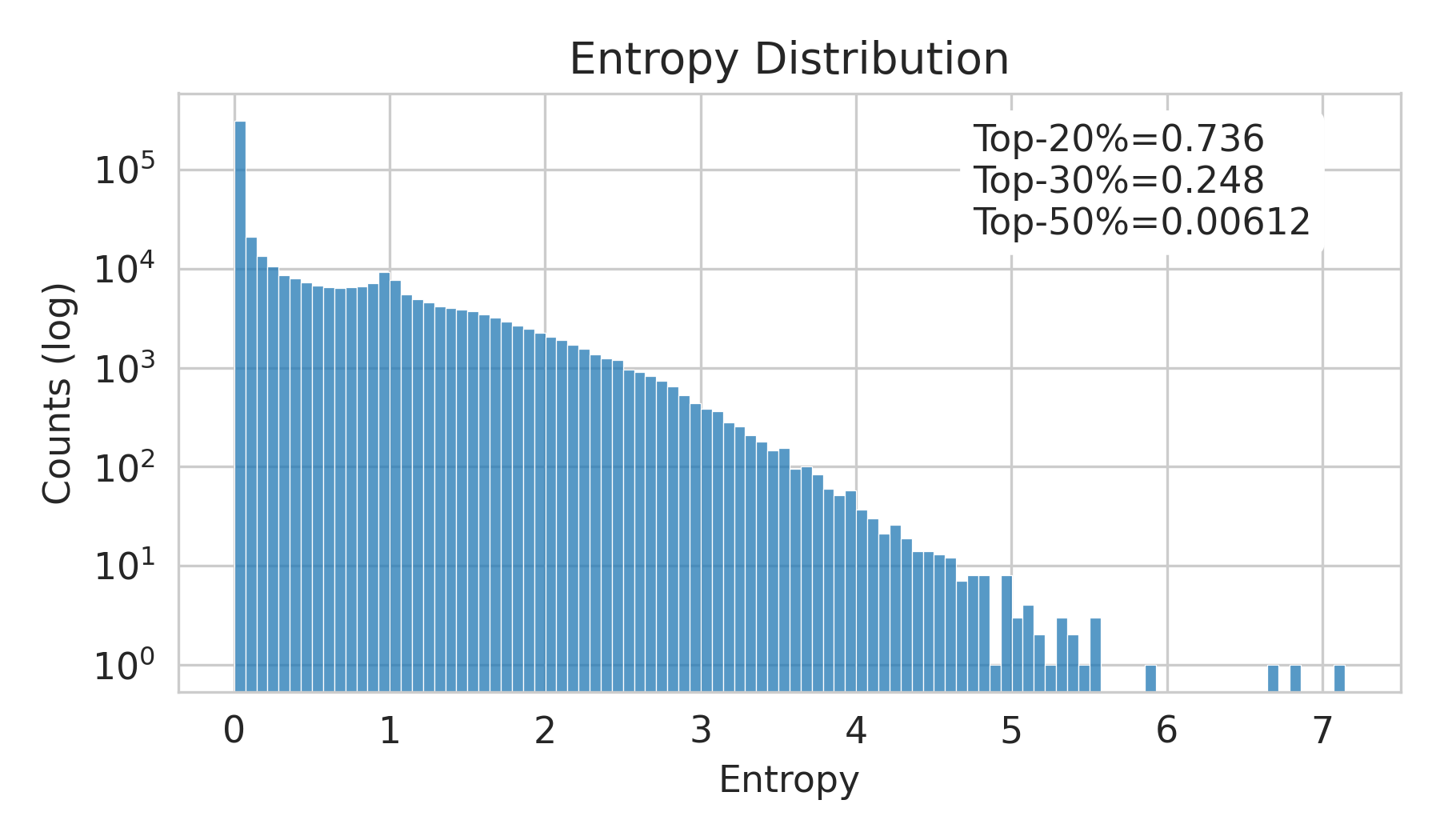}
        \caption{Distribution of entropy values for Qwen-2.5-VL-7B on WeMath, used to identify reasoning-related tokens~\citep{wemath}.}
        \label{fig:entropy-wemath}
    \end{minipage}
\end{figure}

To further investigate the characteristics of perception- and reasoning-related tokens, we analyze the token distribution of Qwen-2.5-VL-7B on both perception and reasoning benchmarks. Specifically, we report the entropy distribution (Figures~\ref{fig:entropy-hallubench}, \ref{fig:entropy-wemath}) and the image log-probability difference distribution (Figures~\ref{fig:logp-diff-hallubench}, \ref{fig:logp-diff-wemath}) on \textit{HallusionBench}~\citep{hallusionbench} and \textit{WeMath}~\citep{wemath}, respectively. For each case, we further highlight the values corresponding to the top 20\%, 30\%, and 50\% of tokens, ranked by their entropy or log-probability differences.

Our observations are as follows:

\ding{182} Perception-related benchmark~\citep{hallusionbench}.
High-entropy tokens exhibit large variations, with more than 50\% of tokens exceeding $0.89$. In contrast, the image log-probability difference of perception-related tokens is relatively stable: more than 70\% of tokens are less than $0.916$. This indicates that a small set of perception-related tokens is consistently stable and plays a critical role in capturing visual cues, suggesting that focusing on these tokens effectively enhances the model's perceptual ability.

\ding{183} Reasoning-related benchmark~\citep{wemath}.
In reasoning tasks, the distribution shows the opposite trend. Perception-related tokens present large variations, with more than 50\% exceeding $0.946$. Meanwhile, high-entropy tokens remain relatively stable, with over 70\% less than $0.248$. This implies that a small subset of reasoning-related tokens is more robust and can effectively capture the key reasoning process, highlighting their importance in enhancing the model's reasoning capability.

These results demonstrate that perception and reasoning rely on different types of stable tokens: perception emphasizes stability in a small number of visually sensitive tokens, while reasoning relies on the robustness of high-entropy tokens. This contrast validates our token re-weighting strategy that explicitly leverages both perception- and reasoning-related tokens.

\section{Perception-Reasoning Interdependence}
\begin{figure}[htbp]
    \centering
    \begin{minipage}{0.45\textwidth}
        \centering
        \includegraphics[width=\textwidth]{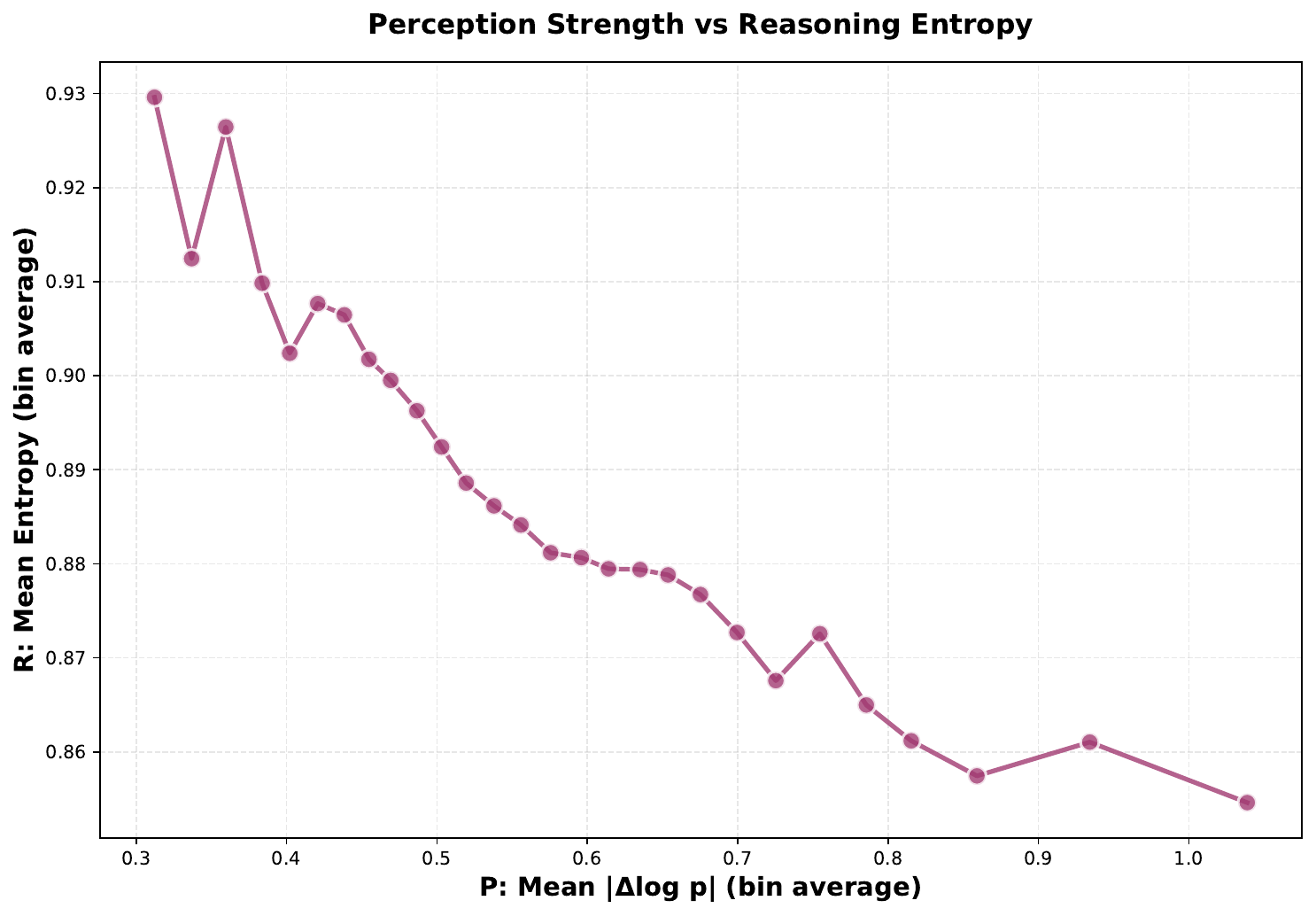}
        \caption{Relationship between (1) reasoning uncertainty: the entropy value of reasoning tokens and (2) perceptron strength: the logp diff between perception tokens for each response over Geo3K training set, response sampled from Qwen-2.5-VL 7B.}
        \label{fig:relation-training}
    \end{minipage}
    \hfill
    \begin{minipage}{0.45\textwidth}
        \centering
        \includegraphics[width=\textwidth]{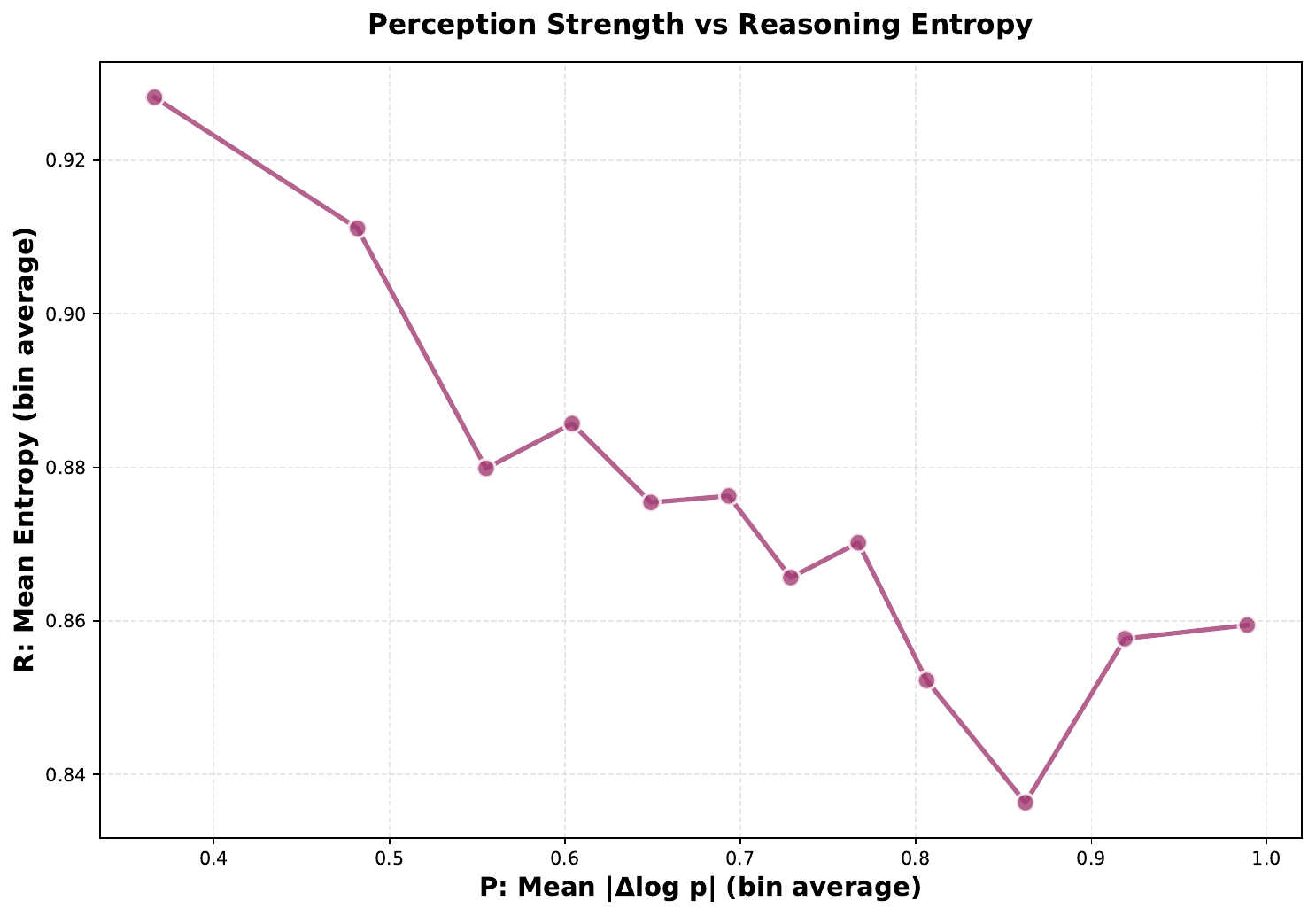}
        \caption{Relationship between (1) reasoning uncertainty: the entropy value of reasoning tokens and (2) perceptron strength: the logp diff between perception tokens for each response over Geo3K validation set, response sampled from Qwen-2.5-VL 7B.}
        \label{fig:relation-validation}
    \end{minipage}
\end{figure}

In this section, we demonstrate the interdependence between perception and reasoning tokens. Specifically, we visualize the relationship between reasoning uncertainty and perception strength over the training and validation set of Geo3K (Figure~\ref{fig:relation-training} \& Figure~\ref{fig:relation-validation}), where we can observe a strong push-pull dynamic relationship between the reasoning and perception. 

\section{Perception Token Identification with Various Criteria}

Identifying perception-related tokens is a key component of Token Reweighting.
Since perception relevance is not directly observable, we investigate several feasible proxy criteria derived from the model’s predicted distributions.
Specifically, we compare the following four alternatives:
\emph{probability difference} (prob-diff), \emph{log-probability difference} (logp-diff), \emph{entropy difference} (entropy-diff), and \emph{attention scores} to image features.

\textbf{Proxy criteria.}
Given a generated token $o_t$, conditioned on either the image $\mathrm{I}$ or an empty image placeholder $\varnothing$, we consider:
(i) prob-diff, measuring absolute probability change;
(ii) logp-diff, measuring relative probability change;
(iii) entropy-diff, measuring distribution-level uncertainty change; and
(iv) attention scores, which directly reflect visual grounding but are computationally expensive to store during rollout and thus used only for analysis.

\begin{figure}[t]
\centering
\includegraphics[width=\textwidth]{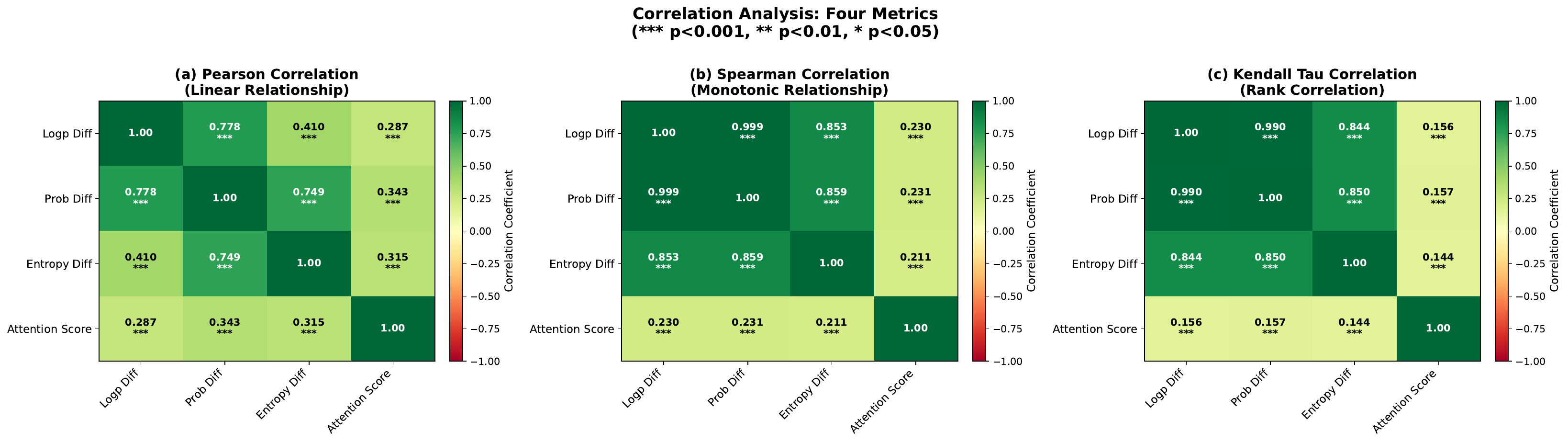}
\caption{Correlation heatmaps between different image token selection strategies over the validation set.}
\label{correlation_heatmaps_geo3k}
\end{figure}

\textbf{Visual grounding fidelity.}
We first evaluate how well each proxy captures visual grounding by measuring its rank correlation with attention scores to image tokens.
Results on Geo3K, WeMath, and HallusionBench are reported in
Figure~\ref{correlation_heatmaps_geo3k} $\to$ Figure~\ref{violin_comparison_extended_new_wemath}.
Among all probability-based proxies, \emph{prob-diff} exhibits the strongest correlation with attention scores, indicating that it best reflects the \emph{absolute influence} of the image on individual token predictions.
In contrast, entropy-diff shows the weakest correlation, as it measures global distribution changes rather than token-level visual dependence.

\textbf{Information-theoretic significance.}
While prob-diff captures visual influence effectively, it treats equal-magnitude probability changes as equally important, regardless of their relative scale.
Entropy-diff, on the other hand, directly measures changes in model uncertainty and is information-theoretically well grounded, but lacks token-level localization.
These observations reveal a fundamental trade-off between accurately capturing \emph{how much the image matters} and \emph{how meaningful the change is}.

\textbf{Log-probability difference as a balanced proxy.}
Logp-diff naturally balances these two aspects.
Empirically, it achieves rank correlations with attention scores close to those of prob-diff, while simultaneously accounting for the relative significance of probability changes.
This makes logp-diff more sensitive to informative but low-probability tokens, which are common in multimodal reasoning.
As a result, logp-diff provides a more faithful token-level estimate of visual relevance.

\textbf{Impact on training performance.}
Table~\ref{tab:proxy_comparison} reports downstream performance when different proxies are used within the ToR framework.
Across HallusionBench, WeMath, and MathVerse, logp-diff consistently yields the largest gains over the GRPO baseline, outperforming both prob-diff and entropy-diff.
This empirical advantage aligns with the above analysis, suggesting that balancing visual grounding fidelity and information-theoretic significance is critical for effective perception token identification.

\textbf{Summary.}
Overall, prob-diff excels at capturing absolute visual influence but lacks information-theoretic sensitivity, while entropy-diff is theoretically principled yet poorly localized.
Logp-diff offers a balanced alternative, preserving strong visual grounding signals while properly weighting the significance of probability changes.
This balance makes logp-diff the most effective and robust proxy for identifying perception-related tokens in Token Reweighting.

\begin{table}[t]
\centering
\caption{\textbf{Comparison of different perception proxies within the ToR framework.}
Performance is reported on HallusionBench, WeMath, and MathVerse. Gains are computed relative to the GRPO baseline.}
\label{tab:proxy_comparison}
\begin{tabular}{lcccc}
\toprule
\textbf{Perception Proxy} & \textbf{HallusionBench} & \textbf{WeMath} & \textbf{MathVerse} & \textbf{Avg. Gain} \\
\midrule
Baseline (GRPO) 
& 69.8 & 67.4 & 50.8 & -- \\
Prob Diff (best visual grounding) 
& 71.8 (+2.0) & 68.5 (+1.1) & 52.6 (+1.8) & +1.6\% \\
Entropy Diff (best info-theoretic) 
& 70.9 (+1.1) & 67.9 (+0.5) & 51.5 (+0.7) & +0.8\% \\
Logp Diff (best balance) 
& \textbf{72.4} (+2.6) & \textbf{68.9} (+1.5) & \textbf{53.0} (+2.2) & \textbf{+2.1\%} \\
\bottomrule
\end{tabular}
\end{table}

\begin{figure}[t]
\centering
\includegraphics[width=\textwidth]{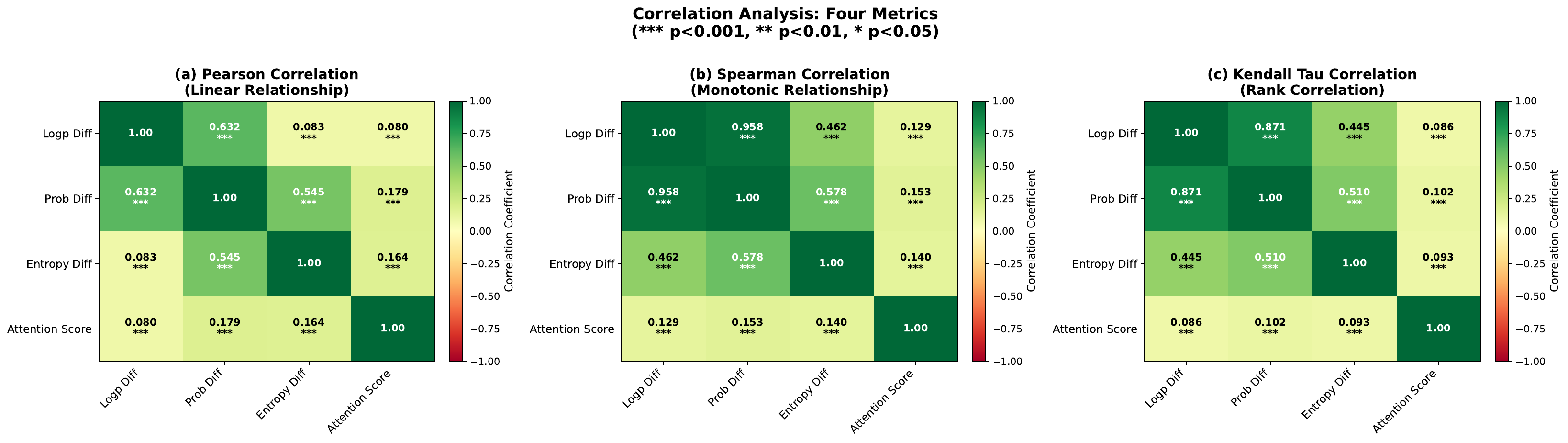}
\caption{Correlation heatmaps between different image token selection strategies over the hallubench benchmark.}
\label{correlation_heatmaps_hallubench}
\end{figure}

\begin{figure}[t]
\centering
\includegraphics[width=\textwidth]{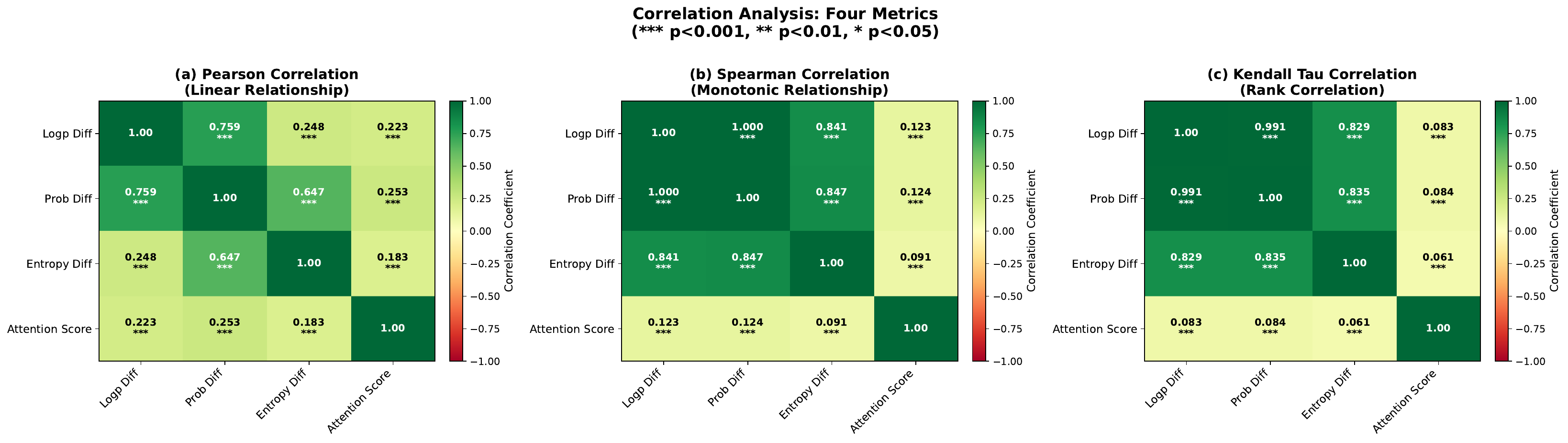}
\caption{Correlation heatmaps between different image token selection strategies over the wemath benchmark.}
\label{correlation_heatmaps_wemath}
\end{figure}

\begin{figure}[t]
\centering
\includegraphics[width=\textwidth]{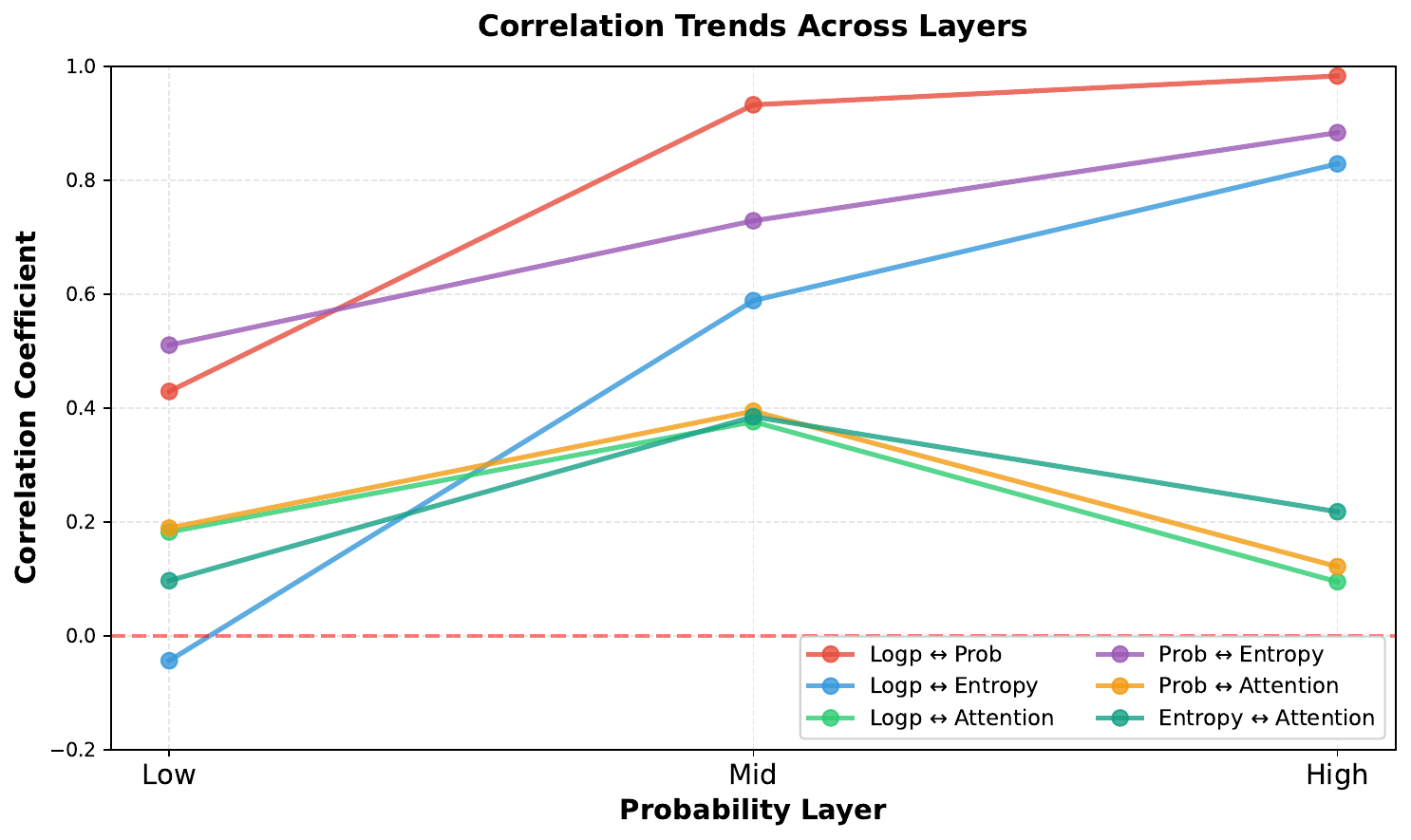}
\caption{Correlation comparison across layers between different image token selection strategies over the validation set.}
\label{correlation_trends_across_layer_geo3k}
\end{figure}

\begin{figure}[t]
\centering
\includegraphics[width=\textwidth]{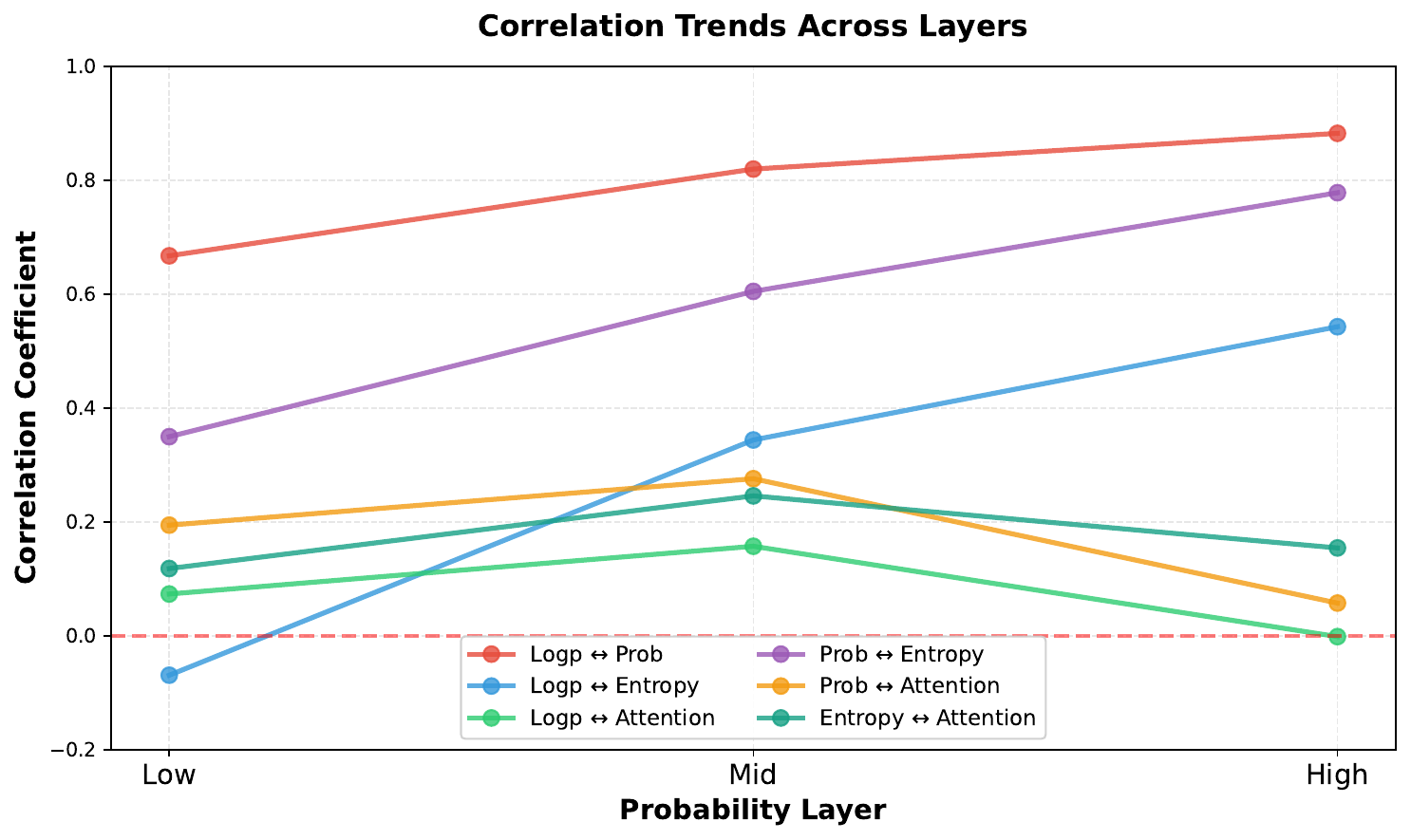}
\caption{Correlation comparison across layers between different image token selection strategies over the hallubench benchmark.}
\label{correlation_trends_across_layer_hallubench}
\end{figure}

\begin{figure}[t]
\centering
\includegraphics[width=\textwidth]{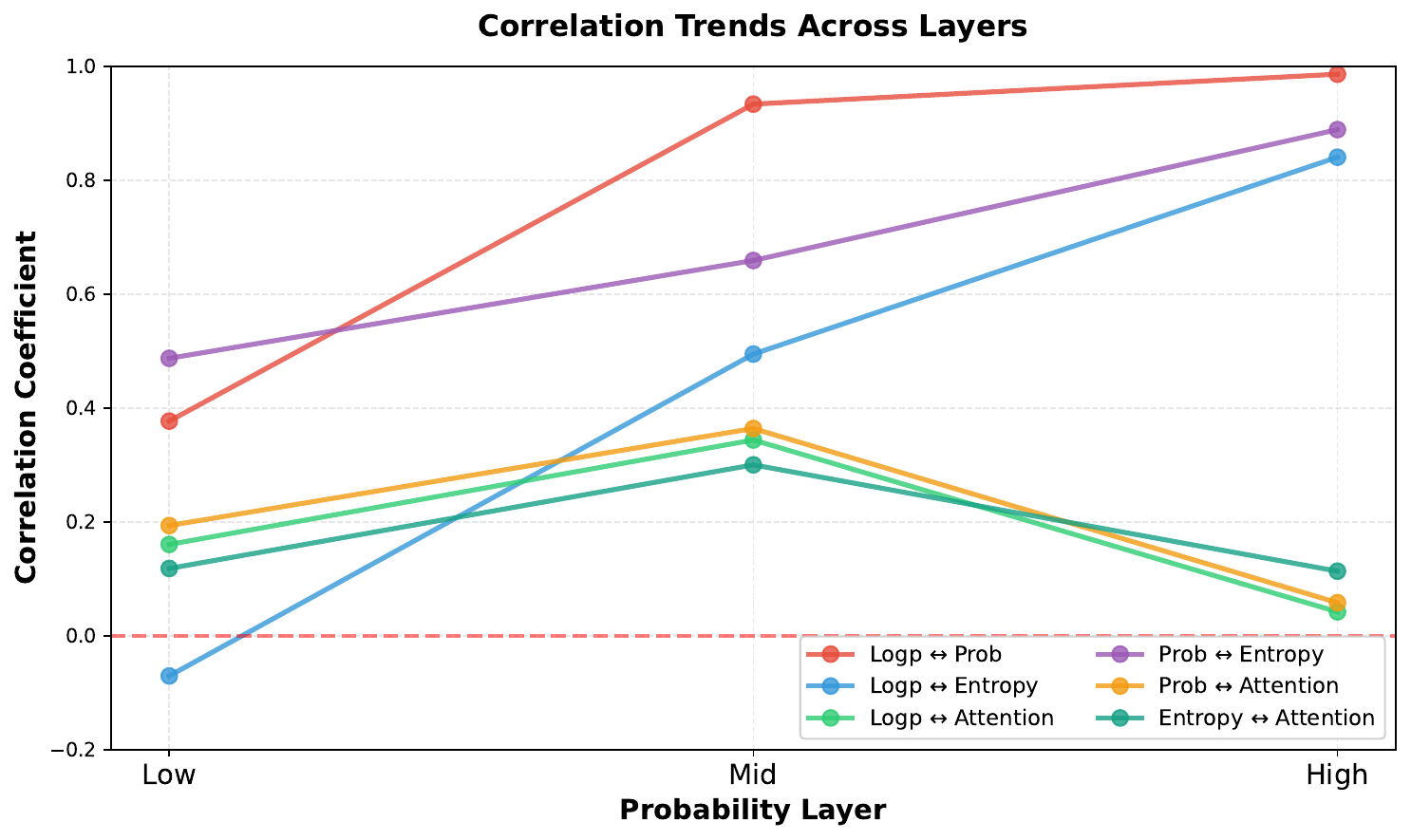}
\caption{Correlation comparison across layers between different image token selection strategies over the wemath benchmark.}
\label{correlation_trends_across_layer_wemath}
\end{figure}

\begin{figure}[t]
\centering
\includegraphics[width=\textwidth]{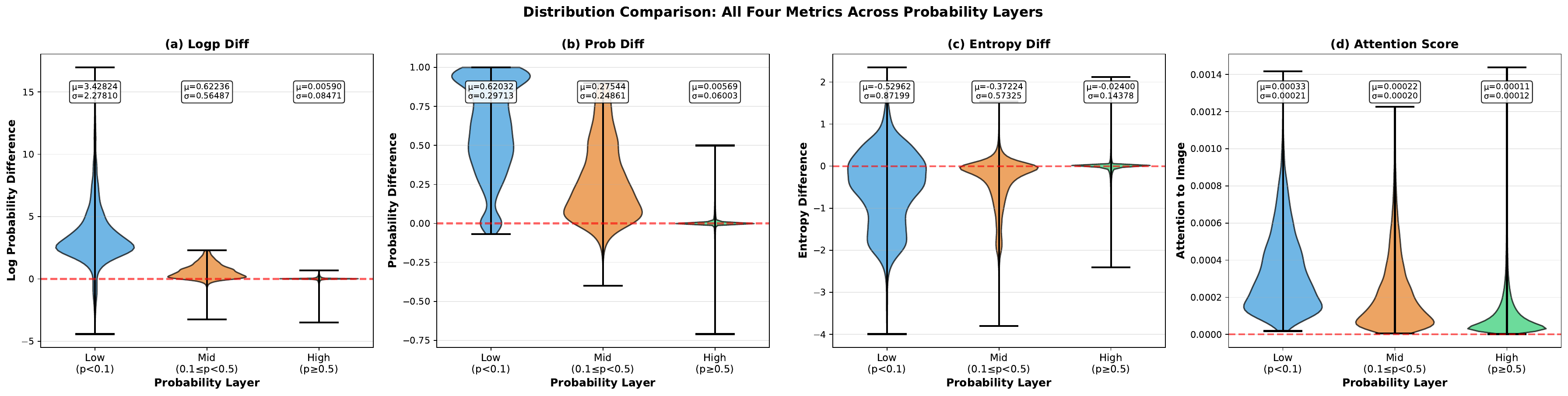}
\caption{violin plot between different image token selection strategies over the validation set.}
\label{violin_comparison_extended_new_geo3k}
\end{figure}

\begin{figure}[t]
\centering
\includegraphics[width=\textwidth]{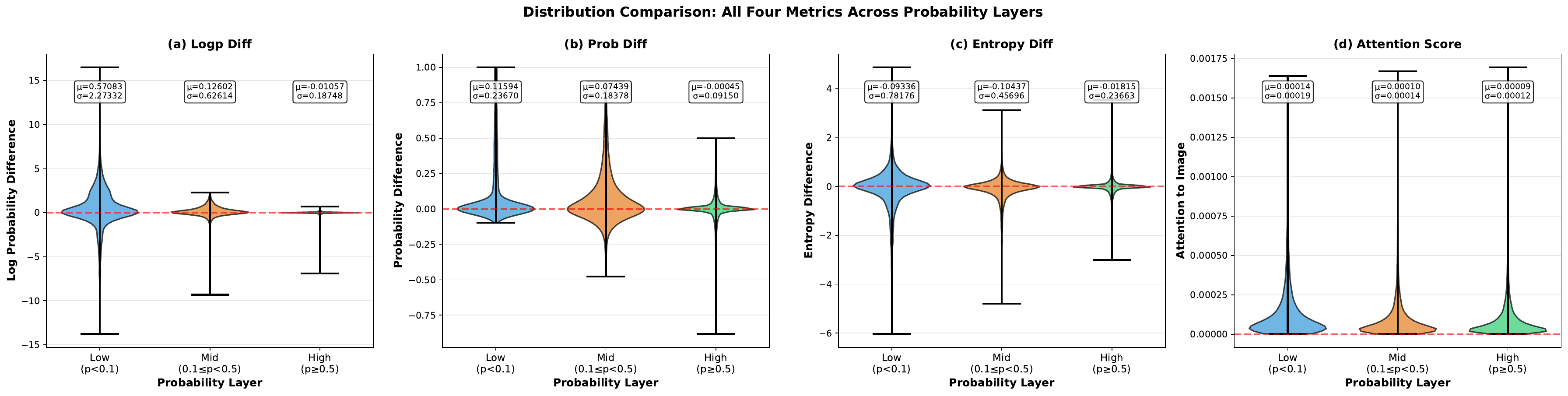}
\caption{violin plot between different image token selection strategies over the Hallubench benchmark.}
\label{violin_comparison_extended_new_hallubench}
\end{figure}

\begin{figure}[t]
\centering
\includegraphics[width=\textwidth]{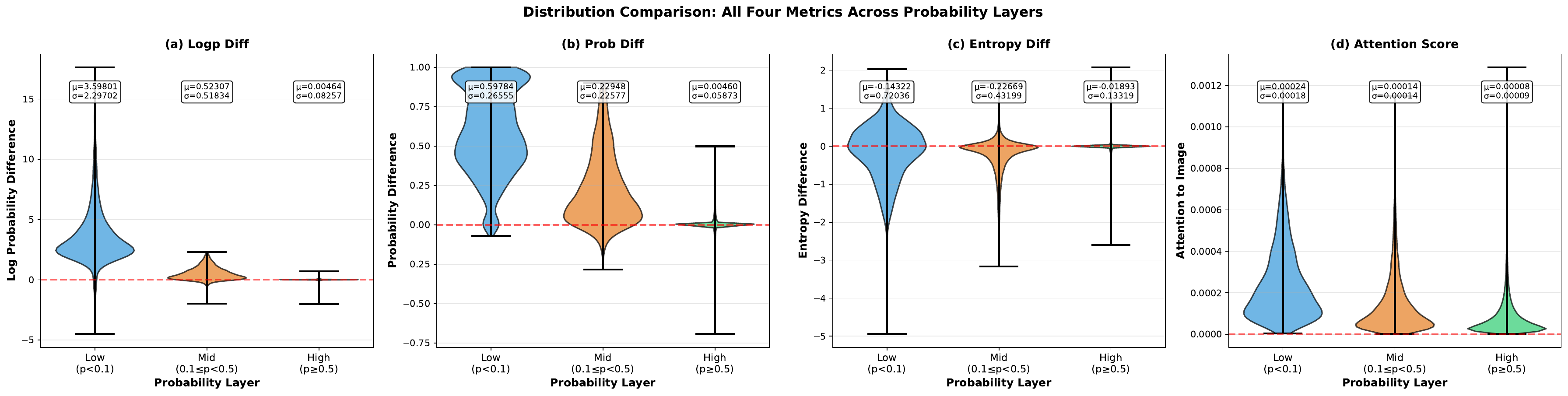}
\caption{violin plot between different image token selection strategies over the wemath benchmark.}
\label{violin_comparison_extended_new_wemath}
\end{figure}

\section{Distribution of Selected Tokens over Various Rollouts.}
\begin{figure}[t]
\centering
\includegraphics[width=\textwidth]{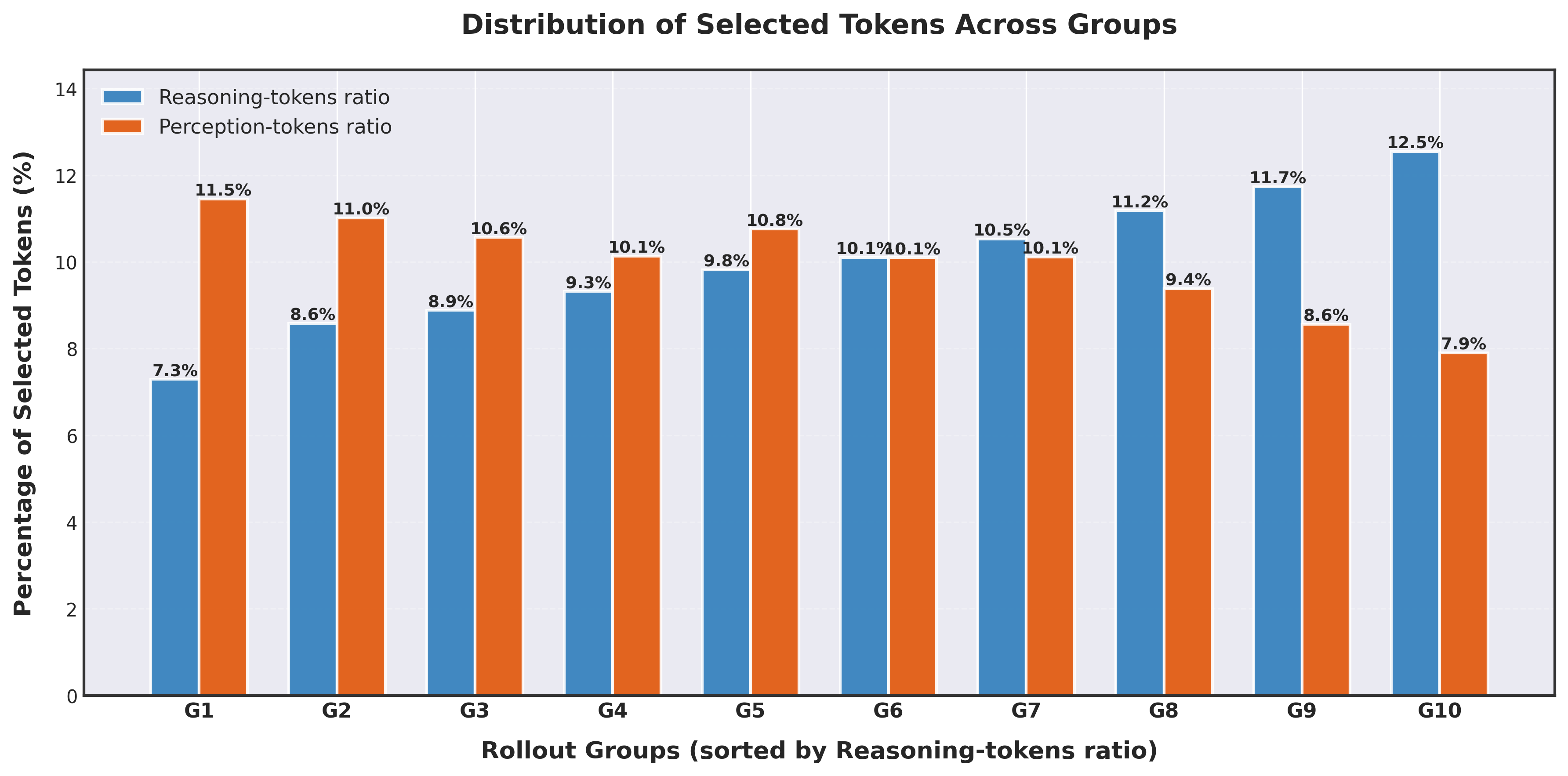}
\caption{Distribution of selected tokens over the sampled rollouts, where we employ the Qwen-VL-2.5 7B model with a batch of 512 samples, each sample generates 16 rollout responses.}
\label{comparison_percentage_chart}
\end{figure}

In this section, we show the distribution of selected tokens over a batch of rollouts as in Figure~\ref{comparison_percentage_chart}. We can find that different rollouts receive a comparable overall amount of optimization, but with different mixtures of tokens: harder groups are optimized more on reasoning, easier groups more on perception, while both token types remain well represented across the rollout batch. 

\end{document}